\documentclass[10pt,twocolumn,letterpaper]{article}

\usepackage[pagenumbers]{cvpr} 

\usepackage{graphicx}
\usepackage{amsmath}
\usepackage{amssymb}
\usepackage{float}
\usepackage{multirow}
\usepackage{makecell}
\usepackage[hypcap=false]{caption}
\usepackage{booktabs}
\usepackage[table]{xcolor}
\usepackage[pagebackref]{hyperref}
\hypersetup{colorlinks,linkcolor={red},citecolor={green},urlcolor={red}}

\definecolor{none_color}{HTML}{FFFFFF}
\definecolor{best_color}{HTML}{FCE5CD}
\definecolor{better_color}{HTML}{DEEDF2}

\DeclareMathOperator*{\minimize}{min}

\newcommand{\pt}{\mathbf{x}} 
\newcommand{\cnlpt}{\pt_c}
\newcommand{\obspt}{\pt_o}
\newcommand{\skelpt}{\pt_{\rm skel}}
\newcommand{\ray}{\mathbf{r}} 
\newcommand{\rayorigin}{\mathbf{o}} 
\newcommand{\raydirection}{\mathbf{d}} 
\newcommand{\rayt}{t}
\newcommand{\depthvalue}{D}
\newcommand{\alphavalue}{A}
\newcommand{\cam}{\mathbf{e}} 

\newcommand{\nroffsetpack}{\Delta \pt}
\newcommand{\loss}{\mathcal{L}}

\newcommand{\coordapp}{\mathit{a}}
\newcommand{\coordbodypose}{\mathit{b}}
\newcommand{\coordcamview}{\mathit{c}}
\newcommand{\appidx}{{\rm idx}_{\rm a}}
\newcommand{\poseidx}{{\rm idx}_{\rm b}}

\newcommand{\viewcam}{{\cam}_{\rm v}}

\newcommand{\cnlvolfunc}{F_c}
\newcommand{\obsvolfunc}{F_o} 
\newcommand{\posecorrectfunc}{P_{\rm pose}}

\newcommand{\motionfield}{T}
\newcommand{\skelmotionfield}{T_{\rm skel}}
\newcommand{\nrmotionfield}{T_{\rm NR}}

\newcommand{\rotbasis}{R}
\newcommand{\transbasis}{\mathbf{t}}

\newcommand{\weightcnl}{w_c}
\newcommand{\weightvolcnl}{W_c}
\newcommand{\weightobs}{w_o}

\newcommand{\weightcnn}{\rm CNN} 
\newcommand{\weightcnnlatent}{\textbf{z}}

\newcommand{\mlp}{\rm MLP} 
\newcommand{\canonicalmlp}{{\rm MLP}_{\appearanceparam}}
\newcommand{\nrmlp}{{\mlp}_{\nrparam}}
\newcommand{\posemlp}{{\mlp}_{\posecorrectparam}}
\newcommand{\posencode}{\gamma} 
\newcommand{\mlpcolor}{\mathbf{c}}
\newcommand{\mlpdensity}{\mathbf{\sigma}}

\newcommand{\bodypose}{\mathit{\mathbf{p}}}
\newcommand{\joints}{J}

\newcommand{\jangles}{\Omega}

\newcommand{\volumerender}{\Gamma} 
\newcommand{\fglikelihood}{f}
\newcommand{\numsamples}{G}

\newcommand{\dataappidx}{s}

\newcommand{\skelparam}{\theta_{\text{skel}}}
\newcommand{\nrparam}{\theta_{\rm NR}} 
\newcommand{\appearanceparam}{\theta_c} 
\newcommand{\posecorrectparam}{\theta_{\rm pose}}

\newcommand{\allparam}{\Theta}
\newcommand{\allparamhumannerf}{\appearanceparam, \skelparam, \nrparam, \posecorrectparam}
\newcommand{\allparamdetail}{\appearanceparam, \skelparam, \posecorrectparam, \allappembedding, \allposeembedding}

\newcommand{\appembedding}{\ell^{\rm app}}
\newcommand{\allappembedding}{L^{\rm app}}
\newcommand{\poseembedding}{\ell^{\rm pose}}
\newcommand{\allposeembedding}{L^{\rm pose}}
\newcommand{\hquad}{\hspace{0.5em}} 
\newcommand{\papertitle}{PersonNeRF}

\begin{document}

\title{\papertitle\ : \\ Personalized Reconstruction from Photo Collections}
\author{Chung-Yi Weng$^{1}$ \hquad Pratul P. Srinivasan$^{2}$ \hquad Brian Curless$^{1,2}$ \hquad Ira Kemelmacher-Shlizerman$^{1,2}$ \\ 
$^1$University of Washington \quad $^2$Google Research \\
\small{\url{https://grail.cs.washington.edu/projects/personnerf/}}
}

\twocolumn[{
\renewcommand\twocolumn[1][]{#1}
\maketitle
    \begin{center}
    \includegraphics[width=\textwidth]{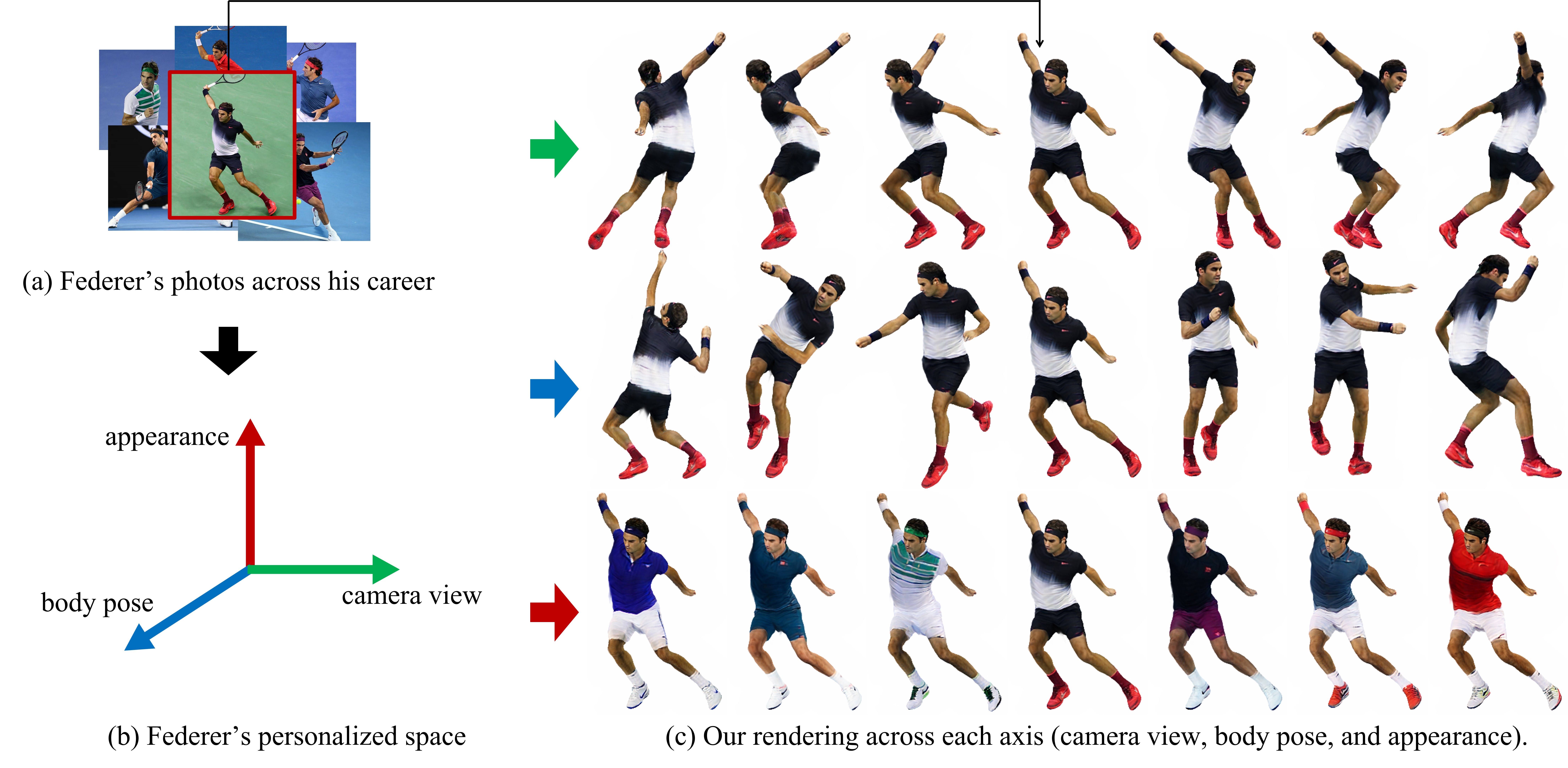}
    \end{center}  
    \vspace*{-10px}
    \captionof{figure}{Given a photo collection of a subject (\eg, Roger Federer) (a), our method \papertitle\ builds a space of the subject spanned by camera view, body pose, and appearance (b). \papertitle\ enables traversing this space and exploring unobserved combinations of these attributes (c). Here we render novel views (top row), various body poses (middle row), and different appearances (bottom row) by traversing the corresponding axes. Among all of the renderings shown here, only the central images of rows correspond to a photo actually observed in the training data (corresponding input photo marked with a red square). \textit{Photo credits to Getty Images.}}
    \label{fig:teaser}
}
\vspace{10px}
]

\begin{abstract}

We present \papertitle\, a method that takes a collection of photos of a subject (\eg Roger Federer) captured across multiple years with arbitrary body poses and appearances, and enables rendering the subject with arbitrary novel combinations of viewpoint, body pose, and appearance. \papertitle\ builds a customized neural volumetric 3D model of the subject that is able to render an entire space spanned by camera viewpoint, body pose, and appearance. A central challenge in this task is dealing with sparse observations; a given body pose is likely only observed by a single viewpoint with a single appearance, and a given appearance is only observed under a handful of different body poses. We address this issue by recovering a canonical T-pose neural volumetric representation of the subject that allows for changing appearance across different observations, but uses a shared pose-dependent motion field across all observations. We demonstrate that this approach, along with regularization of the recovered volumetric geometry to encourage smoothness, is able to recover a model that renders compelling images from novel combinations of viewpoint, pose, and appearance from these challenging unstructured photo collections, outperforming prior work for free-viewpoint human rendering. 

\end{abstract}


\section{Introduction}
\label{sec:intro}

We present a method for transforming an unstructured personal photo collection, containing images spanning multiple years with different outfits, appearances, and body poses, into a 3D representation of the subject. Our system, which we call \papertitle\, enables us to render the subject under novel unobserved combinations of camera viewpoint, body pose, and appearance.

Free-viewpoint rendering from unstructured photos is a particularly challenging task because a photo collection can contain images at different times where the subject has different clothing and appearance. Furthermore,  we only have access to a handful of images for each appearance, so it is unlikely that all regions of the body would be well-observed for any given appearance. In addition, any given body pose is likely observed from just a single or very few camera viewpoints. 

We address this challenging scenario of sparse viewpoint and pose observations with changing appearance by modeling a single canonical-pose neural volumetric representation that uses a shared motion weight field to describe how the canonical volume deforms with changes in body pose, all conditioned on appearance-dependent latent vectors. Our key insight is that although the observed body poses have different appearances across the photo collection, they should all be explained by a common motion model since they all come from the same person. Furthermore, although the appearances of a subject can vary across the photo collection, they all share common properties such as symmetry so embedding appearance in a shared latent space can help the model learn useful priors.

To this end, we build our work on top of HumanNeRF \cite{weng_humannerf_2022_cvpr}, which is a state-of-the-art free-viewpoint human rendering approach that requires hundreds of images of a subject without clothing or appearance changes. Along with regularization, we extend HumanNeRF to account for sparse observations as well as enable modeling diverse appearances. Finally, we build an entire personalized space spanned by camera view, body pose, and appearance that allows intuitive exploration of arbitrary novel combinations of these attributes (as shown in Fig. \ref{fig:teaser}). 
\section{Related Work}
\label{sec:related_work}

\paragraph{3D reconstruction from unstructured photos} Reconstructing static scenes from unstructured photo collections is a longstanding research problem in the fields of computer vision and graphics. The seminal Photo Tourism system~\cite{snavely2006photo} applies large-scale structure-from-motion \cite{schonberger2016structure} to tourist photos of famous sites, enabling interactive navigation of the 3D scene. Subsequent works leveraged multi-view stereo~\cite{seitz2006comparison,furukawa2015multi} to increase the 3D reconstruction quality \cite{shan2013visual, agarwal2011building}. Recently, this problem has been revisited with neural rendering \cite{tewari2020state, tewari2022advances, meshry2019neural, li2020crowdsampling, sun2022neural}. In particular, Neural Radiance Fields (NeRFs) \cite{mildenhall2020nerf} have enabled photorealistic view synthesis results of challenging scenes, including tourist sites~\cite{martinbrualla2020nerfw} and even city-scale scenes~\cite{tancik2022block}. In addition to static scenes, unstructured photo collections have been also used to model human faces \cite{Kemelmacher-Shlizerman_2013_ICCV,liang2016head} or even visualize scene changes through time \cite{martin20153d, martin2015time, matzen2014scene}. 

Our method builds on top of NeRF's neural volumetric representation of static scenes, and extends it to model dynamic human bodies from unstructured photo collections.

\paragraph{3D reconstruction of humans} Many early works in image-based rendering~\cite{szeliski2022computer} have addressed the task of rendering novel views of human bodies. These techniques are largely based on view-dependent texture mapping~\cite{debevec1996modeling}, which reprojects observed images into each novel viewpoint using a proxy geometry. The image-based rendering community has explored many geometry proxies for rendering humans, including depth maps \cite{zitnick2004high, kanade1997virtualized}, visual hulls \cite{matusik2000image}, and parametric human models \cite{carranza2003free}. An alternative technique for 3D reconstruction and rendering of humans is to use 3D scanning techniques to recover a signed distance field representation~\cite{curless1996volumetric,  dou2016fusion4d}, and then extract and texture a polygon mesh~\cite{guo2019relightables, collet2015high,martin2018lookingood}. Recently, neural field representations~\cite{neuralfields2022}, have become popular for modeling humans since they are suited for representing surfaces with arbitrary topology. Methods have reconstructed neural field representations of humans from a variety of different inputs, including 3D scans\cite{tiwari2021neural, chen2021snarf, ma2021scale, mihajlovic2021leap, saito2021scanimate}, multi-view RGB observations\cite{peng2021neural,li2022tava, liu2021neural}, RGB-D sequences\cite{dong2022pina}, or monocular videos \cite{weng_humannerf_2022_cvpr, jiang2022neuman}. Our work is most closely related to HumanNeRF~\cite{weng_humannerf_2022_cvpr}, which reconstructs a volumetric neural field from a monocular video of a moving human. We build upon this representation and extend it to enable reconstructing a neural volumetric model from unstructured photo collections with diverse poses and appearances.

\section{Method}
\label{sec:method}

In this section, we first review HumanNeRF~\cite{weng_humannerf_2022_cvpr} (Sec. \ref{subsec:background}), explain how we regularize it to improve reconstruction from sparse inputs (Sec. \ref{subsec:regularization}), and then describe how we model diverse appearances (Sec. \ref{subsec:appearance_modeling} and \ref{subsec:optimization}). Finally, we describe how we build a personalized space to support intuitive exploration (Sec. \ref{subsec:building_space}).

\begin{figure*}
  \vspace{-10px}
  \centering
  \includegraphics[width=1.0\textwidth]{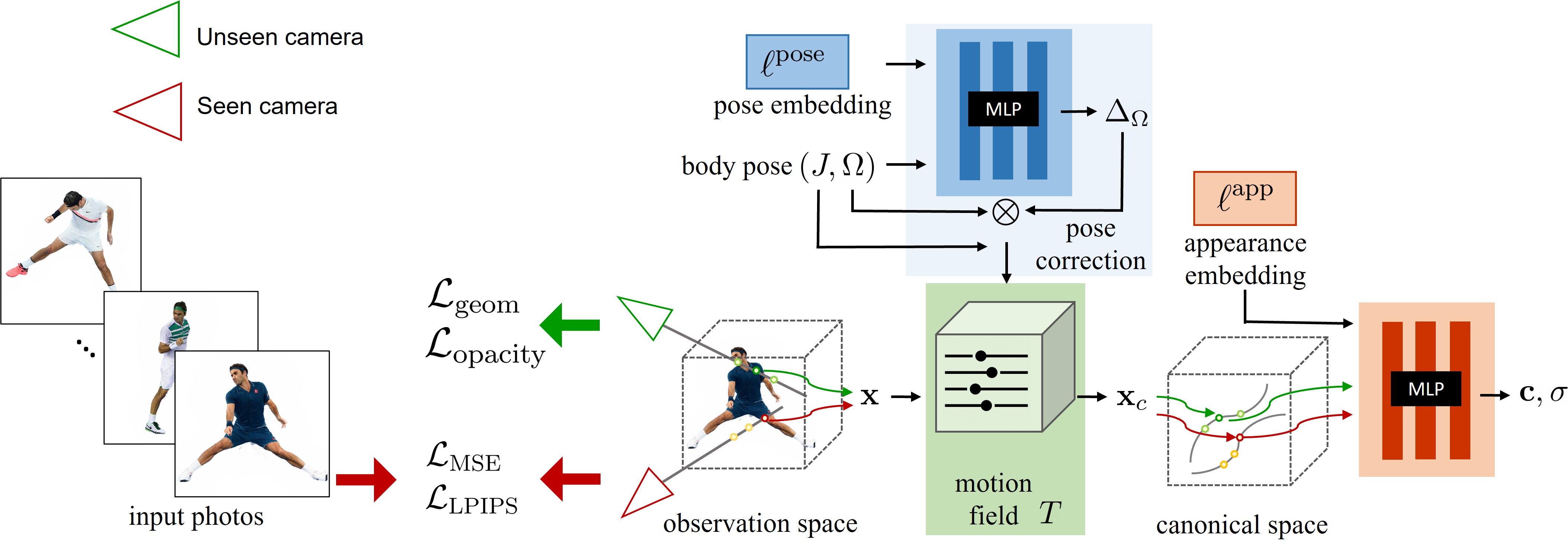}
   \caption{Given an input personal photo collection, our method  optimizes for a canonical volume that can render diverse appearances. We represent the canonical volume with an MLP conditioned on an appearance embedding, and use a shared pose-dependent motion field that maps from observation to canonical space. Additionally, we use a pose correction MLP that takes the estimated body pose and a pose embedding and outputs appearance-dependent pose residuals. Finally, to improve rendering quality from sparse observations, we regularize the volumetric representation to have smooth and opaque geometry with $\loss_{\rm geom}$ and $\loss_{\rm opacity}$, which we apply to renderings from uniformly-sampled unobserved camera viewpoints. \textit{Photo credits to Getty Images.}}
\label{fig:overview}
\vspace{-10px}
\end{figure*}

\subsection{Background}
\label{subsec:background}

\paragraph{HumanNeRF} The recently-introduced HumanNeRF method represents a moving person as a canonical volume $\cnlvolfunc$ warped to a body pose $\bodypose$ to produce a volume $F_o$ in  observed space:
\begin{equation}
\label{eq:canonical_observation_mapping}
\obsvolfunc(\pt, \bodypose) = \cnlvolfunc(\motionfield(\pt,\bodypose)),
\end{equation}
where $\motionfield: (\obspt, \bodypose) \rightarrow \cnlpt$ defines a motion field mapping points from observed space back to canonical space, and $\cnlvolfunc: \pt \rightarrow (\mlpcolor, \mlpdensity)$ maps position $\pt$ to color $\mlpcolor$ and density $\mlpdensity$, represented by $\canonicalmlp(\posencode(\pt))$ taking $\posencode(\pt)$, a sinusoidal positional encoding of $\pt$, as input, with parameters $\appearanceparam$.

The motion field $\motionfield$ is further decomposed into skeletal motion $\skelmotionfield$ and non-rigid motion $\nrmotionfield$:
\begin{equation}
\motionfield(\pt,\bodypose) = \skelmotionfield(\pt,\posecorrectfunc(\bodypose)) 
  + \nrmotionfield(\skelpt, \bodypose), 
\end{equation}
where $\skelpt=\skelmotionfield(\pt,\posecorrectfunc(\bodypose))$, $\nrmotionfield$ represented by $\nrmlp$ predicts a non-rigid offset $\nroffsetpack$, and $\posecorrectfunc(\bodypose)$ corrects the body pose $\bodypose=(\joints,\jangles)$ with the residual of joint angles $\Delta_\jangles$ predicted by $\posemlp(\jangles)$ taking joint angles $\jangles$ as input.

The skeletal motion $\skelmotionfield$ maps an observed position to the canonical space, computed as a weighted sum of $K$ motion bases $(\rotbasis_{i}, \transbasis_{i})$:
\begin{equation}
\skelmotionfield(\pt, \bodypose) = \
  \sum_{i=1}^{K}{\weightobs^i(\pt){(\rotbasis_{i}\pt + \transbasis_{i})}},
\label{eq:skel_motion}  
\end{equation}
where $(\rotbasis_{i}, \transbasis_{i})$, explicitly computed from $\bodypose$, indicates the rotation and translation that maps $i$-th bone from observation to canonical space and $\weightobs^i$ is the corresponding weight in observed space.

Each $\weightobs^i$ is approximated using weights $\weightcnl^i$ defined in canonical space:
\begin{equation}
    \weightobs^i(\pt) =  \frac{\weightcnl^i(\rotbasis_{i}\pt + \transbasis_{i})}
  {\sum_{k=1}^{K}{\weightcnl^k(\rotbasis_{k}\pt + \transbasis_{k})}}.
\label{eq:skel_motion_skin}  
\end{equation}

HumanNeRF stores the set of $\{\weightcnl^i(\pt)\}$ and a background class into a single volume grid $\weightvolcnl(\pt)$ with $K+1$ channels, generated by a convolution network $\weightcnn_{\skelparam}$ that takes as input a random (constant) latent code $\weightcnnlatent$.

\paragraph{Volume Rendering} The observed volume $\obsvolfunc$ that produces color $\mlpcolor$ and density $\mlpdensity$ is rendered using the volume rendering
equation \cite{mildenhall2020nerf}. The expected color $\mathbf{C}(\ray)$ of a ray $\ray(\rayt) = \rayorigin + \rayt\raydirection$ with $\numsamples$ samples is computed as:
\begin{equation}
\label{eq:volume_rendering}
\begin{aligned}
    & \mathbf{C}(\ray) = \sum_{i=1}^{\numsamples} (\prod_{j=1}^{i-1} (1 - \alpha_j))\alpha_i\mlpcolor(\pt_i), \\
    & \quad \alpha_i = \fglikelihood(\pt_i)(1 - \exp(-\mlpdensity(\pt_i) \Delta t_i)), 
\end{aligned}
\end{equation}
where $\Delta t_i = \rayt_{i+1} - \rayt_{i}$ is sample interval, and $\fglikelihood(\pt) = \sum_{k=1}^{K}{\weightcnl^k(\rotbasis_{k}\pt + \transbasis_{k})}$ is foreground likelihood.
Finally, HumanNeRF optimizes for network parameters $\allparam = \{\allparamhumannerf\}$ through MSE loss, $\loss_{\rm MSE}$, and LPIPS \cite{zhang2018unreasonable} loss, $\loss_{\rm LPIPS}$, by comparing renderings with inputs.

\subsection{Unseen view regularization}
\label{subsec:regularization}

Although HumanNeRF \cite{weng_humannerf_2022_cvpr} works well given monocular videos, we observe it produces poor results on unstructured photo collections due to insufficient observations: we usually only have a handful of photos of a subject's  outfit ($<$ 25 images in our case) while HumanNeRF relies on videos with a large number of video frames ($>$ 300 frames).

We find HumanNeRF's struggles in our setting for two reasons: (1) its non-rigid motion does not generalize well to novel viewpoints since there are too few pose observations to sufficiently constrain this pose-dependent effect; (2) the reconstructed canonical-pose human body geometry is incorrect due to insufficient viewpoint observations, resulting in inconsistent appearance in rendered novel viewpoints.

We address the first limitation by simply removing the non-rigid component and only use skeletal motion:
\begin{equation}
\motionfield(\pt,\bodypose) = \skelmotionfield(\pt,\posecorrectfunc(\bodypose))
\end{equation}

We address the second limitation by regularizing the body geometry as rendered in novel views. Specifically, inspired by RegNeRF~\cite{Niemeyer2021Regnerf}, we encourage the geometry to be smooth by enforcing a depth smoothness loss on rendered depth maps. We generate novel camera poses by first sampling an angle $\phi$ from a uniform distribution, $\phi \sim U(0, 2\pi)$, and rotate the input camera with $\phi$ around the up vector with respect to the body center.

We render a pixel's depth value by calculating the expected ray termination position, using the same volume rendering weights used to compute the pixel's color (Eq. \ref{eq:volume_rendering}):
\begin{equation}
\label{eq:expected_depth}
\begin{aligned}
    \depthvalue(\ray) = \sum_{i=1}^{\numsamples} (\prod_{j=1}^{i-1} (1 - \alpha_j))\alpha_i\rayt_{i}.
\end{aligned}
\end{equation}

Likewise, we compute a pixel's alpha value as:
\begin{equation}
\label{eq:alpha_rendering}
\begin{aligned}
    \alphavalue(\ray) = \sum_{i=1}^{\numsamples} (\prod_{j=1}^{i-1} (1 - \alpha_j))\alpha_i.
\end{aligned}
\end{equation}

Our proposed depth smoothness loss is formulated as:
\begin{equation}
\label{eq:depth_smoothness}
\begin{aligned}
    \loss_{\rm geom} = \displaystyle\sum_{i,j=1}^{H-1}
    \left( \alphavalue(\ray_{i,j})\alphavalue(\ray_{i,j+1})( \depthvalue(\ray_{i,j})-\depthvalue(\ray_{i,j+1}))\right)^2 \\
    + \left(\alphavalue(\ray_{i,j})\alphavalue(\ray_{i+1,j})( \depthvalue(\ray_{i,j})-\depthvalue(\ray_{i+1,j}))\right)^2.
\end{aligned}
\end{equation}
where the loss is evaluated over patches of size $H$, as we use patch-based ray sampling similar to HumanNeRF.
Note that this loss only penalizes depth discontinuities when the alphas of neighboring points are high, which effectively constrains the loss to points on the surface.

In practice, we find the depth smoothness term improves geometry and rendering but introduces ``haze" artifacts around the subject. This problem arises because the loss encourages small alphas -- all zero alpha would in fact minimize this term -- biasing toward transparent geometry.

To address this problem, we use an opacity loss inspired by Neural Volumes~\cite{lombardi2019neuralvolumes} that encourages binary alphas:
\begin{equation}
\label{eq:opacity_loss}
\begin{aligned}
    \loss_{\rm opacity} = \displaystyle\sum_{i,j}
    &\log(\alphavalue(\ray_{i,j}) + \epsilon) + \\
    &\log(1 - \alphavalue(\ray_{i,j}) + \epsilon) - C,
\end{aligned}
\end{equation}
where $C = \log(\epsilon) + \log(1 + \epsilon)$ to ensure non-negativity.

\begin{table*}[htbp]
\centering
\renewcommand{\arraystretch}{1.35}
\setlength\tabcolsep{7pt}
\begin{tabular}{|c||c|c|c|c|c|c|c|c|c|c|}

\hline
  & \textbf{2009} & \textbf{2012} & \textbf{2013} & \textbf{2014} & \textbf{2015} & \textbf{2016} & \textbf{2017} & \textbf{2018} & \textbf{2019} & \textbf{2020} \\
\hline
\hline
HumanNeRF \cite{weng_humannerf_2022_cvpr} & 70.64 & 80.62 & 75.09 & 73.00 & 93.89 & 83.35 & 82.19 & 69.40 & 67.47 & 73.01 \\
\hline
Our method & 
\cellcolor{none_color} 59.28 & \cellcolor{none_color} 63.92 & \cellcolor{none_color} 68.92 & \cellcolor{none_color} 63.39 & \cellcolor{none_color} 77.36 & \cellcolor{none_color} 71.99 & \cellcolor{none_color} 71.98 & \cellcolor{none_color} 58.38 & \cellcolor{none_color} 58.21 & \cellcolor{none_color} 61.77 \\
\hline

\end{tabular}
\caption{Comparison to related work: FID is computed per dataset (per year). Lower FID score is better. 
}
\label{table:vs_humannerf}
\vspace{-15px}
\end{table*}

\subsection{Appearance modeling}
\label{subsec:appearance_modeling}
We take as input photos of a subject taken at different times; these photos are subdivided into {\em appearance sets} corresponding to photos taken around the same time, i.e., with the same clothing, etc.

When modeling diverse appearances of a subject, we want to achieve two goals: (1) \textbf{appearance consistency}: synthesizing consistent texture in unobserved regions in one appearance set with the help of the others; (2) \textbf{pose consistency}: a motion model that keeps the rendered pose consistent when switching the subject's appearance.

A naive approach is to train a separate network on each appearance set. This approach does not perform well: (1) the canonical MLP sees very few images in the training, resulting in artifacts in unobserved regions, thus degrading appearance consistency (Fig. \ref{fig:appearance_pose_consistency}-(a)); (2) the learned motion weight volume overfits body poses in each (small) appearance set and does not generalize well to the other sets, leading to poor pose consistency (Fig. \ref{fig:appearance_pose_consistency}-(b)).

Instead, we propose to train all photos with different appearances into a single network. Specifically, we enforce the shared canonical appearance $\canonicalmlp$ to be appearance-dependent but optimize for a single, universal motion weight volume $\weightvolcnl$ across all images.  The shared, appearance-conditioned canonical MLP synthesizes consistent textures by generalizing over the full set of images seen in training, while the universal motion weight volume significantly improves pose consistency, as it is trained on the full set of body poses.

To condition the canonical MLP, inspired by Martin-Brualla et al. \cite{martinbrualla2020nerfw}, we adopt the approach of Generative Latent Optimization \cite{bojanowski2017optimizing}, where each appearance set (with index $i$) is bound to a single real-valued appearance embedding vector $\appembedding_{(i)}$. This vector is concatenated with $\posencode(\pt)$ as input to the canonical $\canonicalmlp$. As a result, the canonical volume $\cnlvolfunc$ is appearance-dependent:
\begin{equation}
\cnlvolfunc(\pt, \appembedding_{(i)}) = {\rm MLP}_{\appearanceparam}(\posencode(\pt), \appembedding_{(i)}).
\end{equation}

Similarly, we introduce pose embedding vector $\poseembedding_{(i)}$ to condition the pose correction module on each appearance set and concatenate this vector with $\jangles$ as input to $\posemlp$.

The appearance embeddings $\allappembedding = \{\appembedding_{(i)}\}_{i=1}^{S}$ as well as pose embeddings $\allposeembedding = \{{\poseembedding_{(i)}}\}_{i=1}^{S}$ are optimized alongside other network parameters, where $S$ is the number of appearance sets.

\subsection{Optimization}
\label{subsec:optimization}

\paragraph{Loss function} Our total loss is a combination of the previously-discussed losses:
\begin{equation}
\label{eq:loss_function}
\begin{aligned}
    \loss = \loss_{\rm LPIPS} + \lambda_{1}\loss_{\rm MSE} + \lambda_{2}\loss_{\rm geom} + \lambda_{3}\loss_{\rm opacity}.
\end{aligned}
\end{equation}

\paragraph{Objective} Given input images $\{I_1, I_2, ..., I_N\}$, appearance set indices $\{\dataappidx_1, \dataappidx_2, ..., \dataappidx_N\}$, body poses $\{\bodypose_1, \bodypose_2, ..., \bodypose_N\}$, and cameras $\{\cam_1, \cam_2, ..., \cam_N\}$, we optimize the objective:
\begin{equation}
\label{eq:optim_problem_def}
    \minimize_{\allparam} \sum_{i=1}^N \loss(\volumerender[\cnlvolfunc(\motionfield(\pt, \bodypose_i, \poseembedding_{(\dataappidx_i)}), \appembedding_{(\dataappidx_i)}), \cam_i], I_i),
\end{equation}
where $\loss(\cdot)$ is the loss function and $\volumerender[\cdot]$ is a volume renderer, and we minimize the loss with respect to all network parameters and embedding vectors $\allparam = \{\allparamdetail\}$.

 We shoot rays toward both seen and unseen cameras. $\loss_{\rm LPIPS}$ and $\loss_{\rm MSE}$ are computed from the output of seen cameras, while $\loss_{\rm geom}$ and $\loss_{\rm opacity}$ are applied to renderings of unseen ones. We use $\lambda_1 = 0.2$, $\lambda_2 = 1.0$, and $\lambda_3 = 10.0$. Additionally, we stop the gradient flow through the pose MLP when backpropagating $\loss_{\rm geom}$, as we found it can lead to degenerate pose correction.

\subsection{Building a personalized space}
\label{subsec:building_space}

Once the optimization converges, we use its result to build a personalized space of the subject spanned by camera view, body pose, and appearance.  We allow continuous variation in viewpoint, but restrict body pose and appearance to those that were observed in the set. Every point in the space has a corresponding rendering. 

In practice, the space is defined as a cube with size 1 where the coordinate value ranges from 0 to 1. Our goal is to map a point in that cube to the inputs of the network from which we render the subject.

Specifically, assuming the subject has $N$ body poses and $S$ appearances, we need to perform mapping on coordinates ($\coordapp, \coordbodypose, \coordcamview$) corresponding to position along the axes of appearance, body pose, and camera view, respectively: 

(1) \textbf{Appearances}: we map the value $\coordapp$ to the index of $S$ appearances: $\appidx = \lfloor{\coordapp S}\rfloor$, which was used to retrieve the appearance embedding $\appembedding_{(\appidx)}$ for canonical $\canonicalmlp$.

(2) \textbf{Body pose}: we map the value $\coordbodypose$ to the index of $N$ body poses: $\poseidx = \lfloor{\coordbodypose N}\rfloor$. We get the $\poseidx~$-th body pose $\bodypose$, corresponding to appearance index $\dataappidx_{\poseidx}$. We then take pose embedding $\poseembedding_{(\dataappidx_{\poseidx})}$ as input for pose $\posemlp$.

(3) \textbf{Camera view}: we rotate the camera $\cam_{\poseidx}$ by $\phi = 2\pi c$ around up vector with respect to the body center to get a viewing camera $\viewcam$. 

Finally, we generate a subject rendering corresponding to the position ($\coordapp, \coordbodypose, \coordcamview$) by feeding the appearance embedding $\appembedding_{(\appidx)}$, pose embedding $\poseembedding_{(\dataappidx_{\poseidx})}$, and body pose $\bodypose$ to the network and producing a volume in observation space rendered by the viewing camera $\viewcam$.

\begin{figure*}
  \vspace{-10px}
  \centering
  \includegraphics[width=1.0\textwidth]{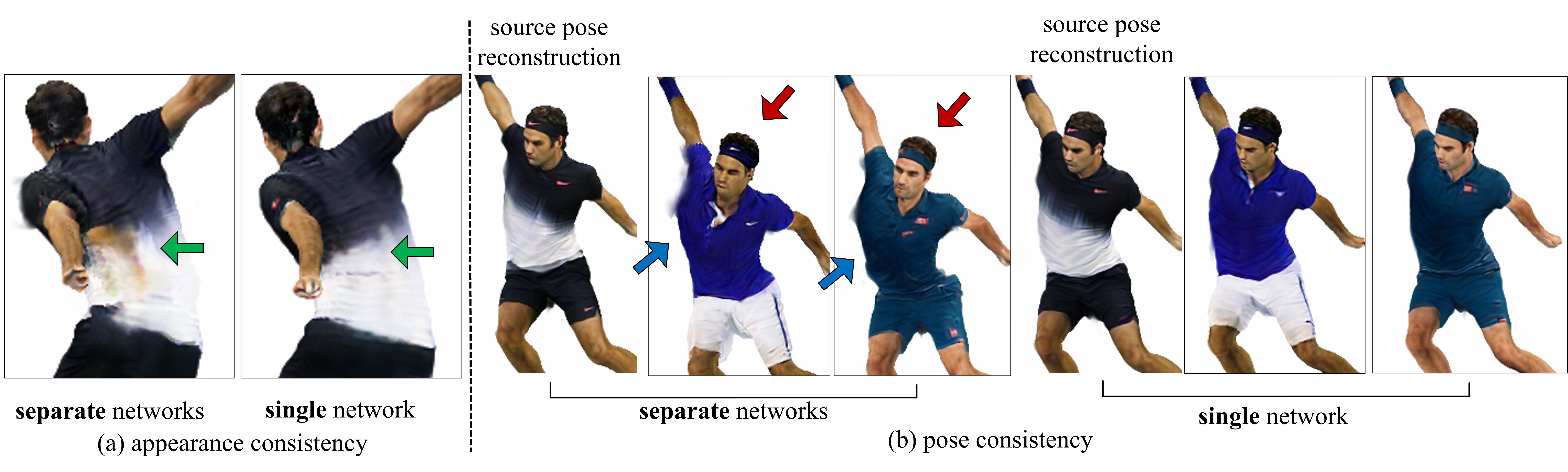}
  \vspace{-15px}
   \caption{(a) \textbf{Appearance consistency}: training all appearance sets with a single network synthesizes higher quality texture for unobserved regions while training with separate networks produces incompatible colors (green arrow). (b) \textbf{Pose consistency}: In comparison to the source pose reconstruction (i.e., the combination of pose and appearance is observed in training), separate-networks training produces unsatisfied results when combining the pose with unseen appearances; the head orientations are different from the input (red arrow) and the bodies are unnaturally distorted (blue arrow). In contrast, single-network optimization enables consistent output.}
\label{fig:appearance_pose_consistency}
\vspace{-10px}
\end{figure*}

\section{Results}
\label{sec:results}

\subsection{Dataset}

In the main paper, we include results on experiments using a photo collection of Roger Federer (more subjects in supplementary material). The Roger Federer dataset contains 10 appearance sets spanning 12 years. We collect photos by searching for a specific game in a particular year (e.g., ``2019 Australian Open Final''). We collected 19 to 24 photos for each game, one per year, and label each set according to the year (\textit{2009}, \textit{2012}, ..., \textit{2020}).

Following \cite{weng_humannerf_2022_cvpr}, we run SPIN \cite{kolotouros2019spin} to estimate body pose and camera pose, automatically segment the subject, and manually correct segmentation errors and 3D body poses with obvious errors. Additionally, for images where the subject is occluded by balls or rackets, we label the regions of occluded objects and omit them during optimization.

\subsection{Implementation details}

We optimize Eq. \ref{eq:optim_problem_def} using the Adam optimizer \cite{kingma2014adam} with hyperparameters $\beta_1 = 0.9$ and $\beta_2 = 0.99$. We set the learning rate to $5 \times 10^{-4}$ for $\appearanceparam$ (the canonical $\mlp$), $\allappembedding$, and $\allposeembedding$ (embedding vectors), and $5 \times 10^{-5}$ for all the others. We sample 128 points along each ray for rendering. The size of embedding vectors of $\appembedding$ and $\poseembedding$ are 256 and 16. We use patch-based ray sampling with 6 patches with size 32x32 for seen cameras and 16 patches with size 8x8 for unseen ones. The optimization takes 200K iterations to converge when training each game with individual networks and takes 600K iterations for all games into a single network. Additionally, we delay pose refinement, geometry regularization, and opacity constraint until after 1K, 1K, and 50K iterations for separate-networks training, and 1K, 10K, and 200K iterations for single-network optimization.

\subsection{Comparison}

\textbf{Baseline } We compare our method with HumanNeRF \cite{weng_humannerf_2022_cvpr}, the state-of-the-art free-viewpoint method on monocular videos. We run experiments on individual datasets (\textit{2009}, \textit{2012}, ..., \textit{2020}). We use the official HumanNeRF implementation with hyperparameters $T_s=2.5K$ and $T_e=5K$ to accommodate the much smaller input dataset size. Because HumanNeRF only can optimize for a single appearance, we do the same in our method. Finally, we train HumanNeRF with 200K iterations, the same number used in our method.

\textbf{Evaluation protocol } As we lack ground truth when evaluating results rendered from unseen views, we adopt Frechet inception distance (FID) \cite{heusel2017gans} for quantitative comparison. For each input image, we rotate the camera in 10-degree increments around the ``up'' vector w.r.t the body center and use these renderings for evaluation.

\begin{figure*}
  \centering
  \includegraphics[width=0.94\textwidth]{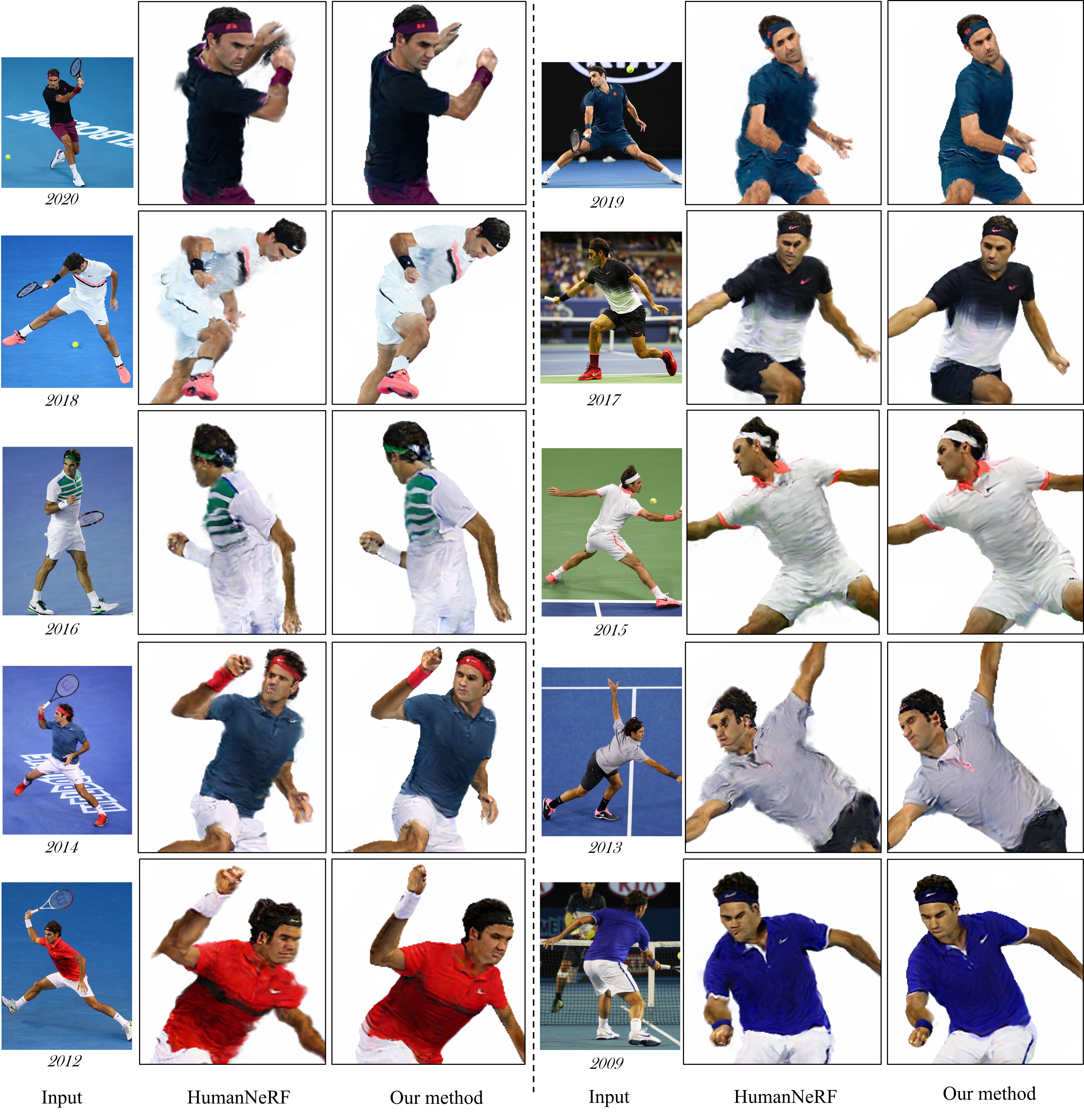}
  \caption{Our method produces more convincing renderings with fewer artifacts than those from HumanNeRF~\cite{weng_humannerf_2022_cvpr}. Note how HumanNeRF produces errors in regions occluded from the input view, while our method produces plausible geometry. \textit{Photo credits to Getty Images.}}
  \label{fig:vs_humannerf}
  \vspace{-15px}
\end{figure*}

\textbf{Results } Quantitatively, as shown in Table \ref{table:vs_humannerf}, our method outperforms HumanNeRF on all datasets by comfortable margins. The performance gain is particularly significant when visualizing the results, as shown in Fig. \ref{fig:vs_humannerf}. Our method is able to create consistent geometry, sharp details, and nice renderings, while HumanNeRF tends to produce irregular shapes, distorted textures, and noisy images, due to insufficient inputs.

\begin{figure}[H]
\centering
\includegraphics[width=\linewidth]{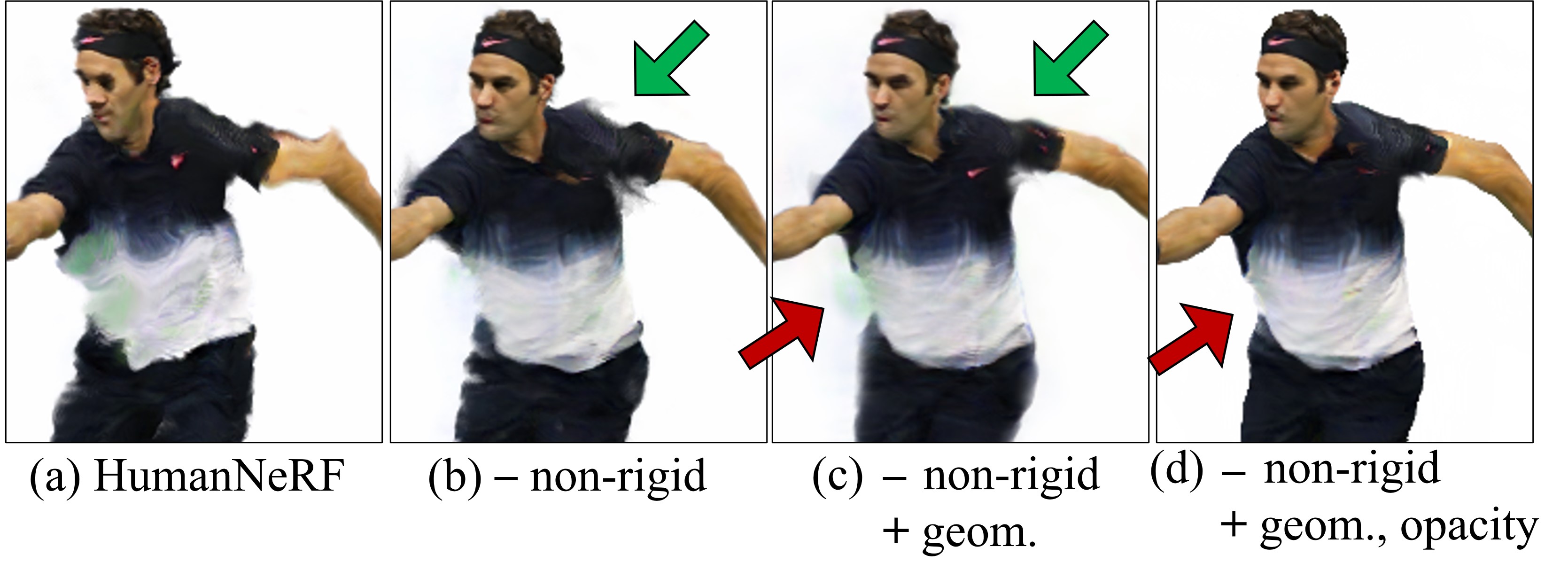}
\vspace{-20px}
\caption{Ablation study. Removing the non-rigid motion component from HumanNeRF significantly improves reconstruction quality. Adding our geometry loss further refines the shape (green arrow) but introduces ``haze'' artifacts (red arrow), which we address with the opacity loss.}
\label{fig:ablation}
\vspace{-10px}
\end{figure}

\textbf{Ablation studies } Fig. \ref{fig:ablation} shows visually how we outperform HumanNeRF by modifying the model and introducing new losses. By removing non-rigid motion, we get a significant quality boost. We further enhance the shape and texture reconstruction with the geometry and opacity losses. Table \ref{table:ablation} quantifies the importance of each element. We get the best performance when including all the refinements.

\begin{table}[H]
\centering
\renewcommand{\arraystretch}{1.0}
\setlength\tabcolsep{20pt}
\begin{tabular}{|l|c|}
\hline
 & FID $\downarrow$  \\
\hline 
\hline
HumanNeRF \cite{weng_humannerf_2022_cvpr}& 76.87 \\
\hline
\makecell[cl]{Ours \quad $-$ non-rigid} & 71.75  \\
\hline
\makecell[cl]{Ours \quad $-$ non-rigid \\ \quad\quad\quad $+$ geometry} & 76.84  \\
\hline
\makecell[cl]{Ours \quad $-$ non-rigid \\ \quad\quad\quad $+$ opacity} & 67.01  \\
\hline
\makecell[cl]{Ours \quad $-$ non-rigid \\ \quad\quad\quad $+$ geometry, opacity } & \textbf{65.52 } \\
\hline
\end{tabular}
\caption{Ablation: average FID (lower is better) over 10 datasets. 
}
\label{table:ablation}
\vspace{-10px}
\end{table}

\textbf{Appearance and pose consistence } Fig. \ref{fig:appearance_pose_consistency} illustrates the benefit of training all images with a single network. In contrast to individually trained networks, Fig. \ref{fig:appearance_pose_consistency}-(a) illustrates it can synthesize compatible textures for unobserved regions as a result of better generalization, thus maintaining appearance consistency; Fig. \ref{fig:appearance_pose_consistency}-(b) demonstrates the unified network is able to keep the rendered body pose persistent across different appearances, thanks to the shared motion weight volume, hence guaranteeing pose consistency.

\begin{figure*}
  \centering
  \includegraphics[width=0.98\textwidth]{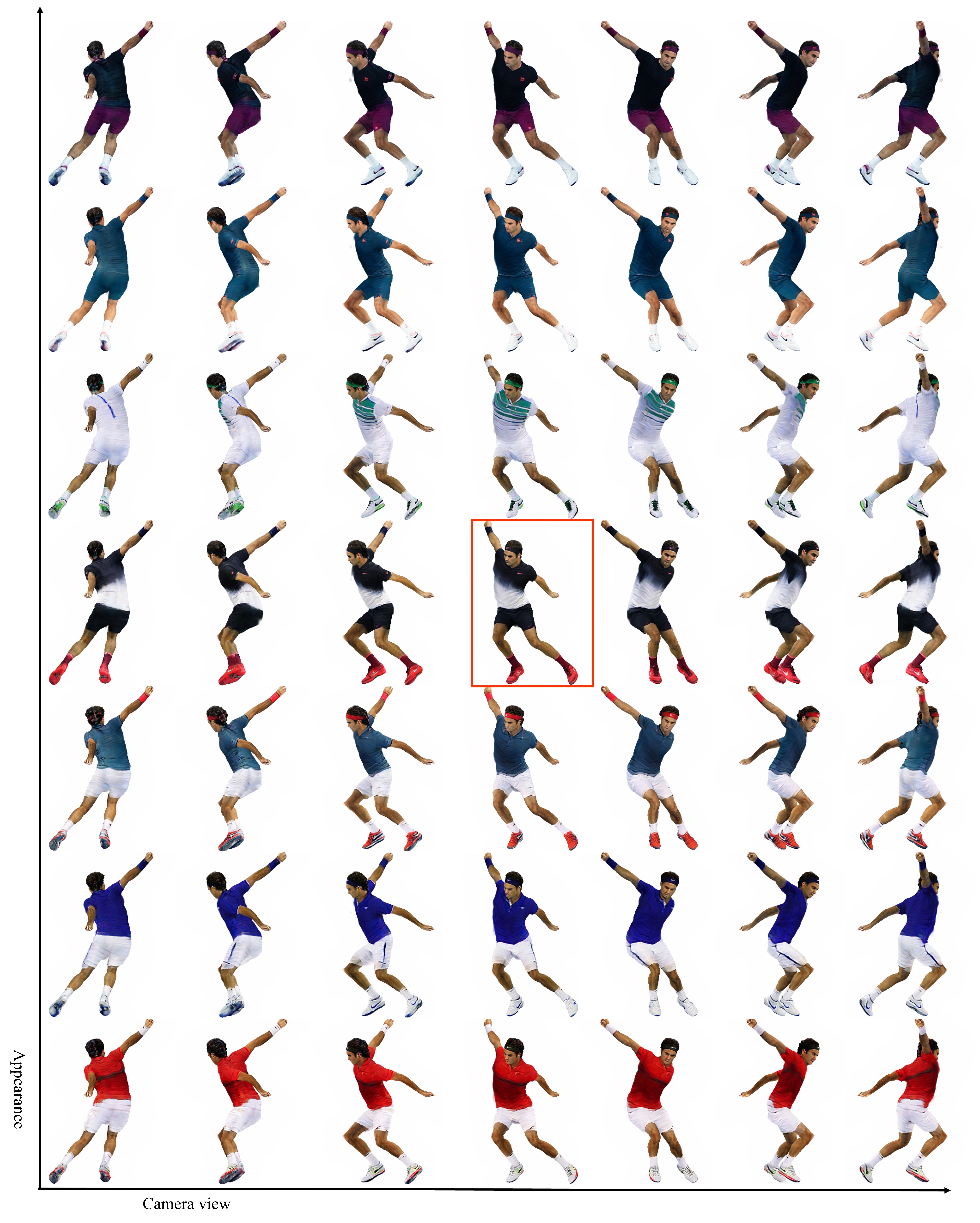}
  \caption{The visualization of the (appearance, camera view) plane of the reconstructed Federer space. Note that only the image in the red square was directly observed in the input data.}
  \label{fig:fed_view_appearance_plane}
\end{figure*}

\textbf{Visualization of Federer space }  In Fig. \ref{fig:fed_view_appearance_plane}, we visualize the rebuilt Federer space by keeping the body pose fixed and rendering dense samples in the camera-appearance plane starting from one photo. In this case, only a single image (the one with a red square) is directly observed, showing how sparse observations we have to rebuild the space. The renderings are sharp and with few artifacts, and the appearance and pose consistency are well-maintained.

\section{Discussion}
\label{sec:discussion}

\textbf{Limitations } Our work builds upon HumanNeRF to account for sparse inputs and diverse appearance. While it is effective in this challenging scenario, it inherits some of HumanNeRF's limitations such as its reliance on the initialized poses, its assumption of relatively diffuse lighting, and its requirement for manual human segmentation. Additionally, since human body pose estimators typically fail on images with heavily-occluded bodies, we can only use input photos that view the full body. 

\textbf{Societal impact } In this work, we aim to faithfully produce images of a person with the capability of just rendering unseen views and switching appearance within their own set of appearances. The work does not intend to create motions and animations that didn't happen. While we focus in the paper only on one person and show more examples in the supplementary material, it is important  to validate in future work that the method scales to a wide range of subjects.

\textbf{Conclusion } We have presented \papertitle\, allowing rendering a human subject with arbitrary novel combinations of body pose, camera view, and appearance from an unstructured photo collection. Our method enables exploring these combinations by traversing a reconstructed space spanned by these attributes and demonstrates high-quality and consistent results across novel views and unobserved appearances. 

\textbf{Acknowledgement:} We thank David Salesin and Jon Barron for their valuable feedback. This project is a tribute from the first author, a die-hard tennis fan, to Novak, Rafa, Roger, and Serena. He feels blessed to have lived in their era and wishes it would never come to an end. This work was funded by the UW Reality Lab, Meta, Google, Oppo, and Amazon.

{\small
\bibliographystyle{ieee_fullname}
\bibliography{egbib}
}

\clearpage
\section*{Supplementary Material}

\appendix

\section{Network Architecture}

\begin{figure}[htbp]
  \centering
  \includegraphics[width=\linewidth]{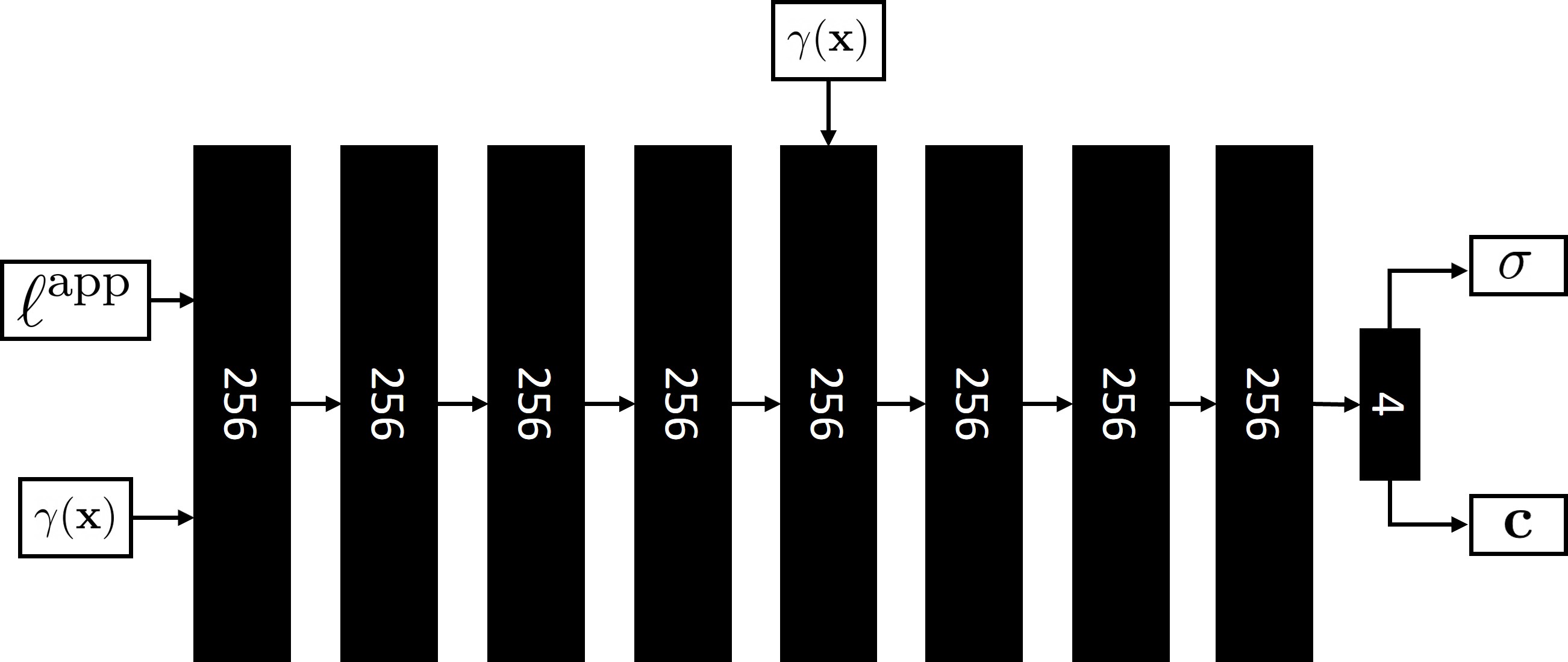}
  \caption{Canonical $\mlp$ netwrok. We use an 8-layer MLP with width=256 that takes as input positional encoding $\posencode$ of position $\pt$ and appearance embedding vector $\appembedding$ with dimension=256. The network outputs color $\mlpcolor$ and density $\mlpdensity$, following the design of NeRF \cite{mildenhall2020nerf}. There is a skip connection that concatenates $\posencode(\pt)$ to the fifth layer. We use ReLU activations after each fully connected layer. For the output layer, we use a ReLU activation for the density value $\mlpdensity$ to ensure non-negativity and a \textit{sigmoid} activation for the color $\mlpcolor$ to constrain values between 0 and 1.}
  \label{fig:mlp_canonical}
\end{figure}

\begin{figure}[htbp]
  \centering
  \includegraphics[width=0.6\linewidth]{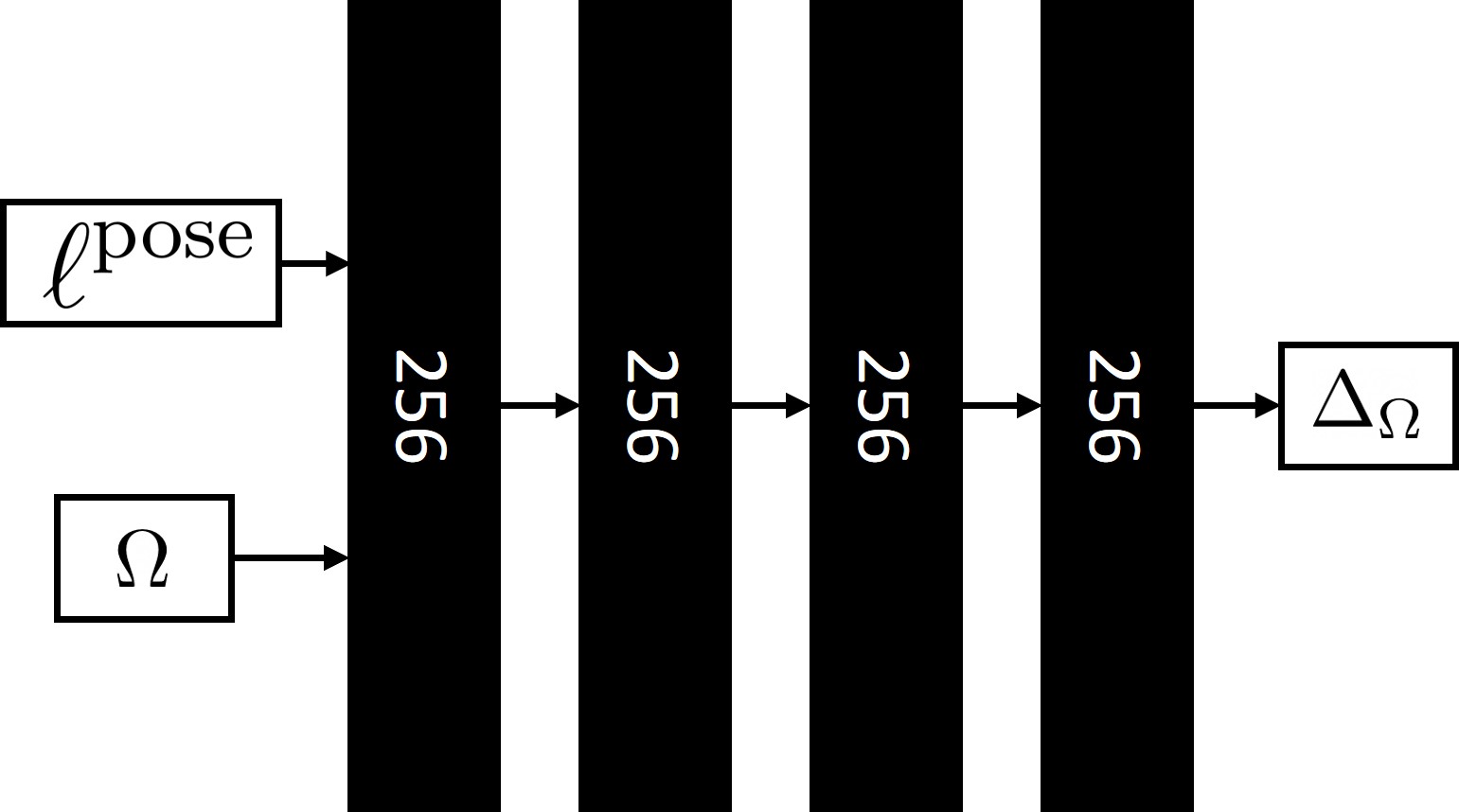}
  \caption{Pose correction $\mlp$ network. We use a 4-layer MLP with width=256 that takes as input joint angles $\jangles$ and a pose embedding vector $\poseembedding$ with dimension=16. The network produces the residuals of joint angles that are added back to the input pose to refine the body pose prediction.}
  \label{fig:mlp_pose_correct}
\end{figure}

Fig. \ref{fig:mlp_canonical} and Fig. \ref{fig:mlp_pose_correct} show the network design of our canonical MLP and pose correction MLP. Specifically, we provide the details of how we incorporate appearance embedding $\appembedding$ as well as pose embedding $\poseembedding$ vectors into the corresponding networks.

\section{Additional Results}

\begin{figure*}
  \centering
  \includegraphics[width=1.0\textwidth]{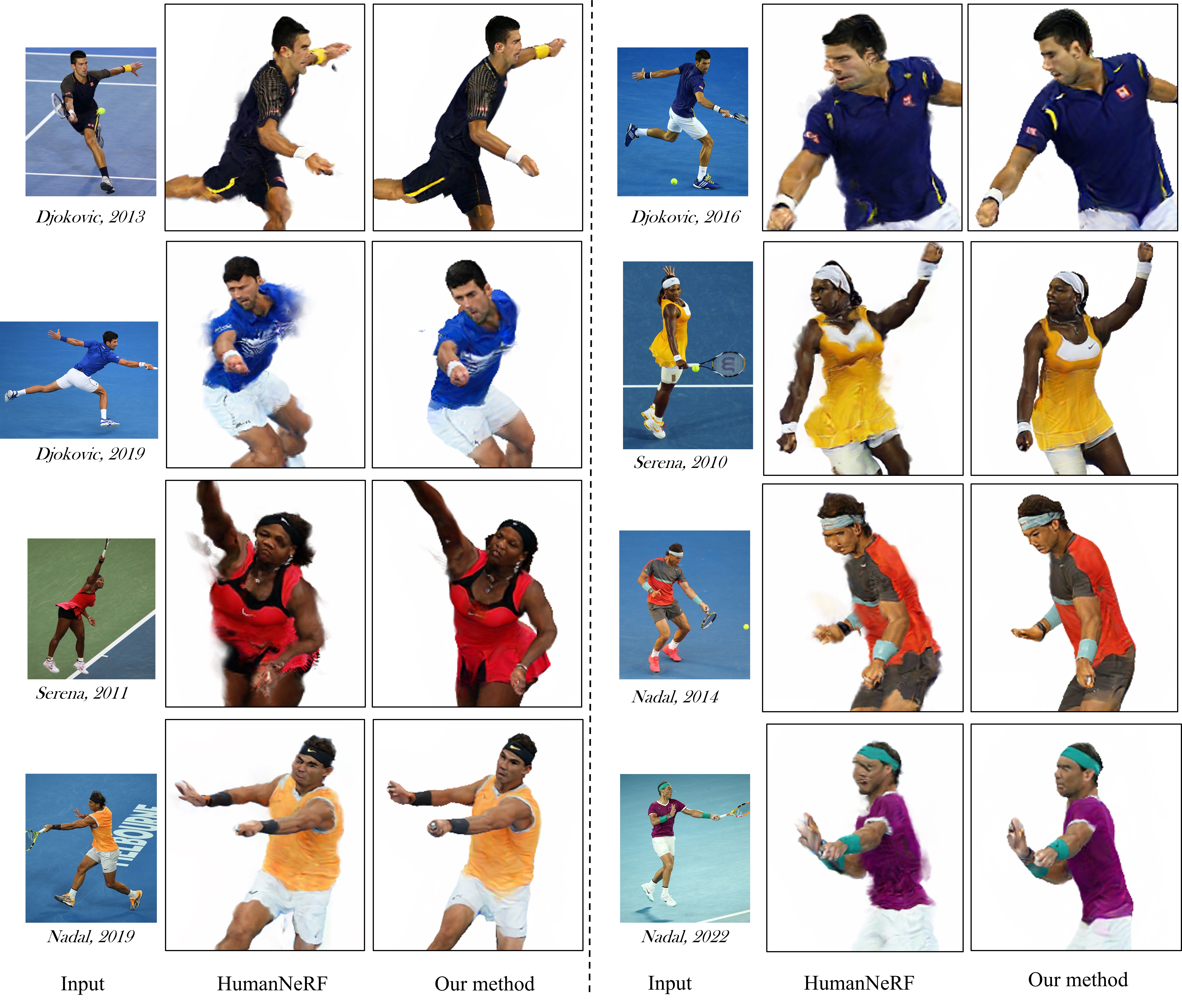}
  \caption{Visual comparisons to HumanNeRF~\cite{weng_humannerf_2022_cvpr} on Novak Djokovic, Serena Willams, and Rafael Nadal. \textit{Photo credits to Getty Images.}}
  \label{fig:vs_humannerf_more}
  \vspace{-15px}
\end{figure*}

In addition to Roger Federer, we demonstrate our method on a wide variety of subjects that cover different genders and skin tones. In particular, we show results on three tennis athletes, Novak Djokovic, Serena Williams, and Rafael Nadal where each has three appearance sets in the datasets we collected. We present quantitative results in FID in Table \ref{table:vs_humannerf_more} and visually compare them with HumanNeRF \cite{weng_humannerf_2022_cvpr} in Fig. \ref{fig:vs_humannerf_more}. The quality improvement over the related work is similar to the case of Roger Federer.

\begin{table*}[htbp]
\centering
\renewcommand{\arraystretch}{1.35}
\setlength\tabcolsep{7pt}
\begin{tabular}{|c || c | c | c || c | c | c || c | c | c|}

\hline
  \multirow{2}{*}{} &  \multicolumn{3}{c||}{\textbf{Novak Djokovic}} & \multicolumn{3}{c||}{\textbf{Serena Willams}} & \multicolumn{3}{c|}{\textbf{Rafael Nadal}} \\ 
\cline{2-10}
  & \textbf{2013} & \textbf{2016} & \textbf{2019} & \textbf{2009} & \textbf{2010} & \textbf{2011} & \textbf{2014} & \textbf{2019} & \textbf{2022} \\
\hline
HumanNeRF \cite{weng_humannerf_2022_cvpr} & 87.07 & 62.01 & 64.17 & 104.23 & 100.52 & 113.41 & 90.04 & 64.95 & 76.68 \\
\hline
Our method & 
81.38 & 57.14  & 58.74 & 87.81 & 90.70 & 85.17 & 80.95 & 62.75 & 61.71 \\
\hline

\end{tabular}
\caption{Comparison to related work: FID is computed per subject per year. Lower FID score is better.
}
\label{table:vs_humannerf_more}
\end{table*}

\section{More Visualizations of Personalized Space}

\begin{figure*}
  \centering
  \includegraphics[width=0.98\textwidth]{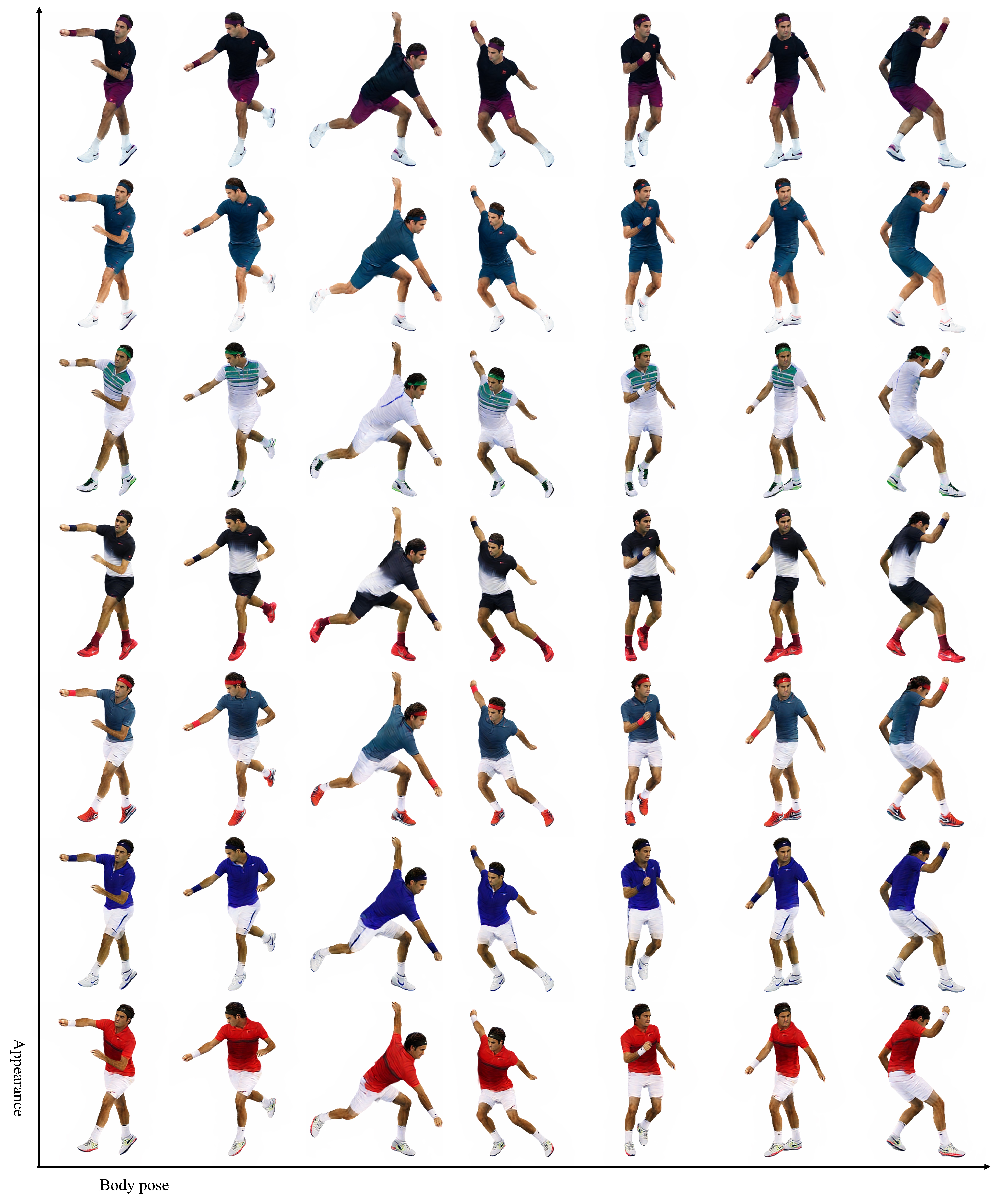}
  \caption{Visualization of the (appearance, body pose) plane of the reconstructed space of Roger Federer. }
  \label{fig:fed_pose_appearance_plane}
\end{figure*}

\begin{figure*}
  \centering
  \includegraphics[width=0.98\textwidth]{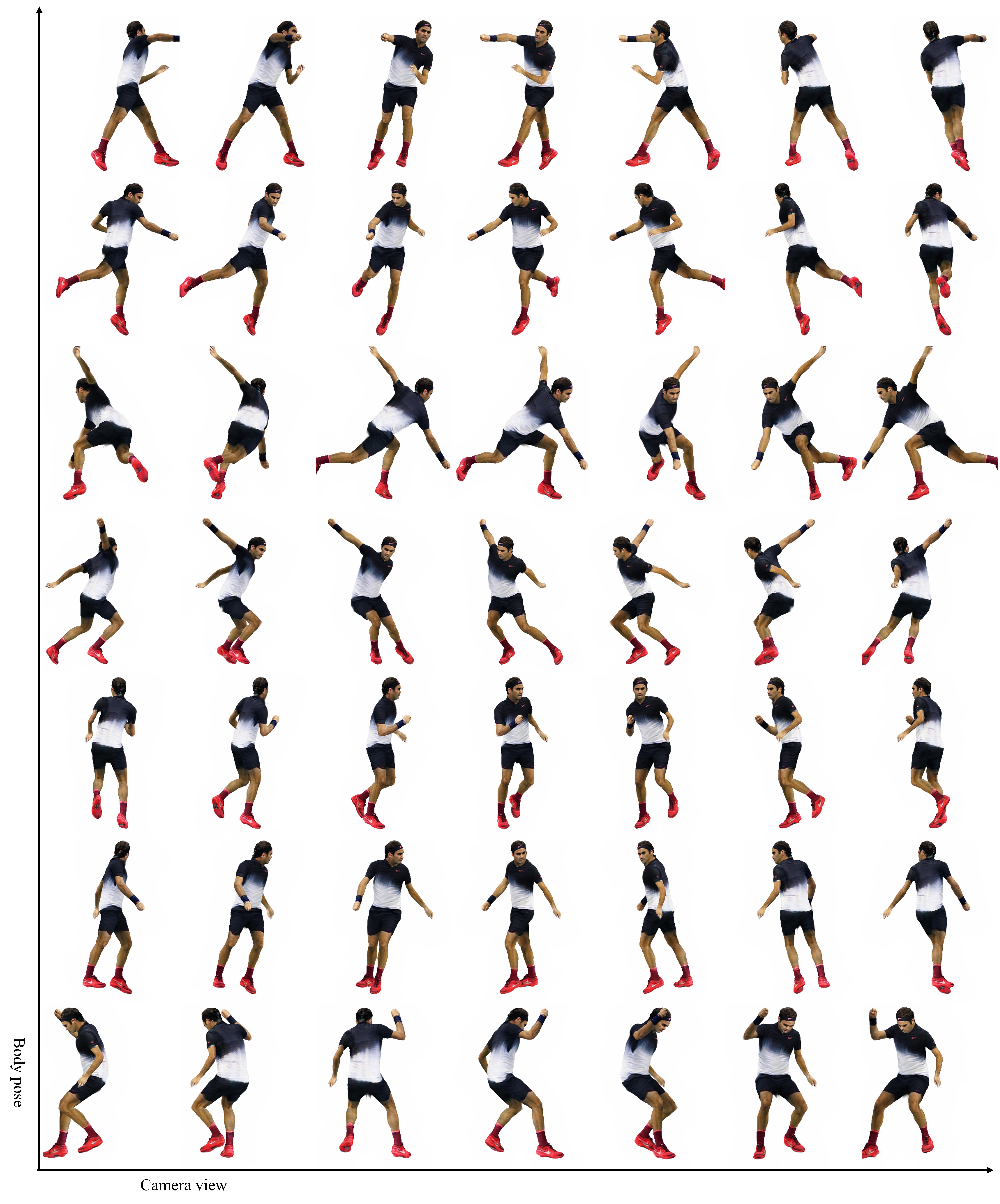}
  \caption{Visualization of the (body pose, camera view) plane of the reconstructed space of Roger Federer. }
  \label{fig:fed_pos_view_plane}
\end{figure*}

In the paper, we show a visualization of (appearance, camera view) plane of the reconstructed space of Roger Federer. Here we show the other two planes, (appearance, body pose) plane in Fig. \ref{fig:fed_pose_appearance_plane} and (body pose, camera view) plane in Fig. \ref{fig:fed_pos_view_plane} where we keep the camera view and appearance fixed, respectively.

\begin{figure*}
  \centering
  \includegraphics[width=\textwidth]{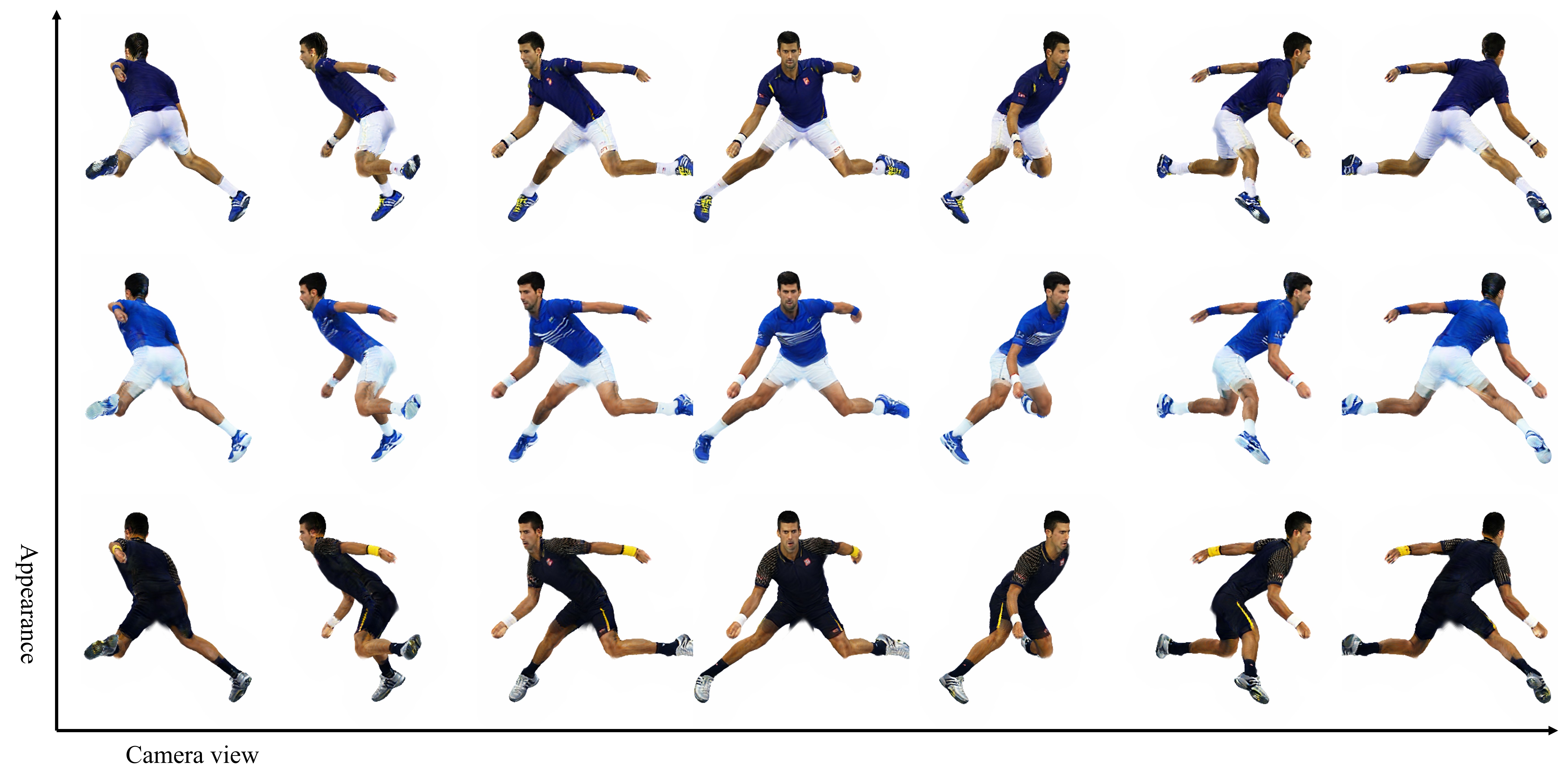}
  \caption{Visualization of the (appearance, camera view) plane of the reconstructed space of Novak Djokovic.}
  \label{fig:djoko_view_appearance_plane}
\end{figure*}

\begin{figure*}
  \centering
  \includegraphics[width=\textwidth]{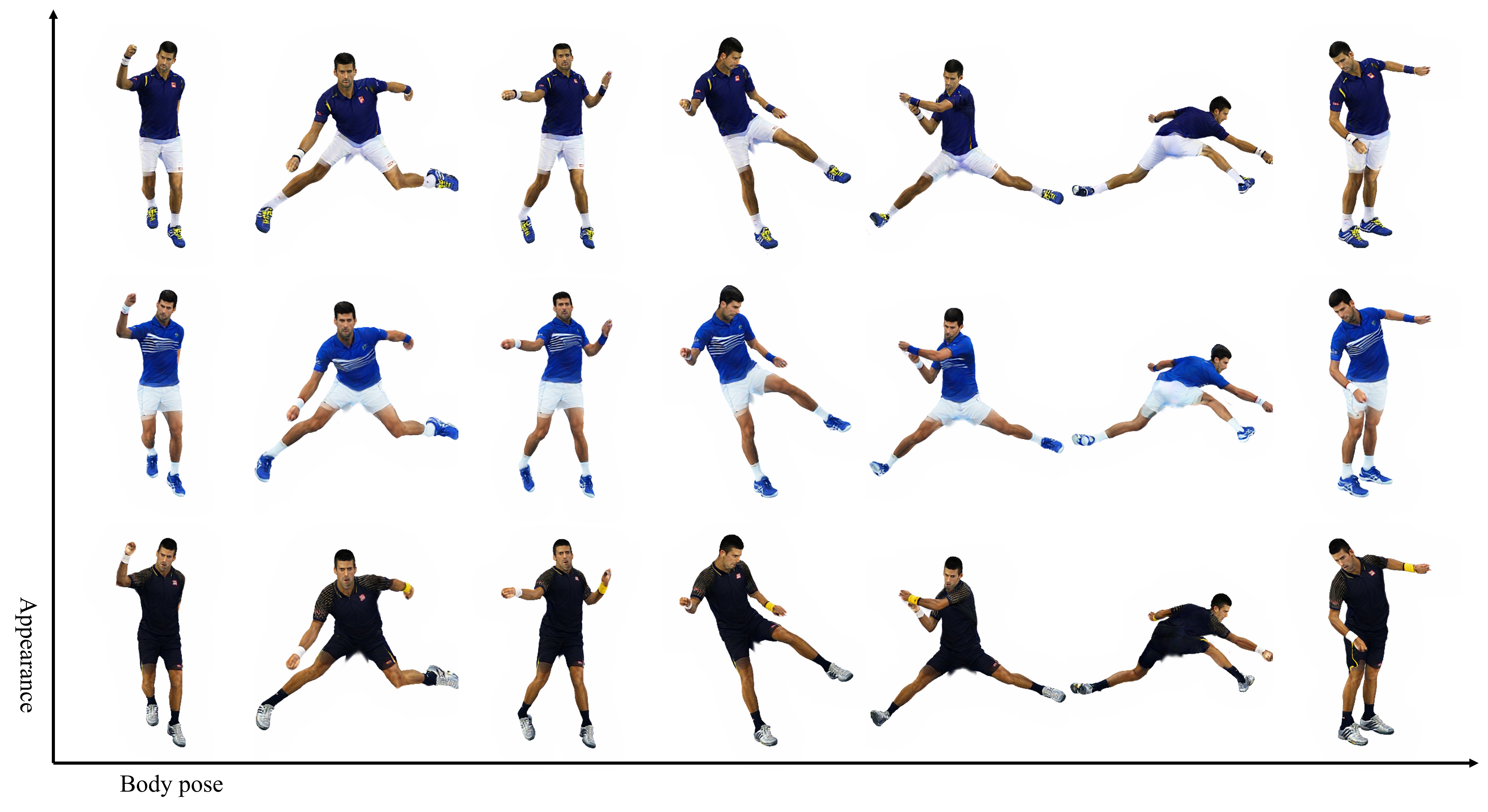}
  \caption{Visualization of the (appearance, body pose) plane of the reconstructed space of Novak Djokovic. }
  \label{fig:djoko_pose_appearance_plane}
\end{figure*}

\begin{figure*}
  \centering
  \includegraphics[width=\textwidth]{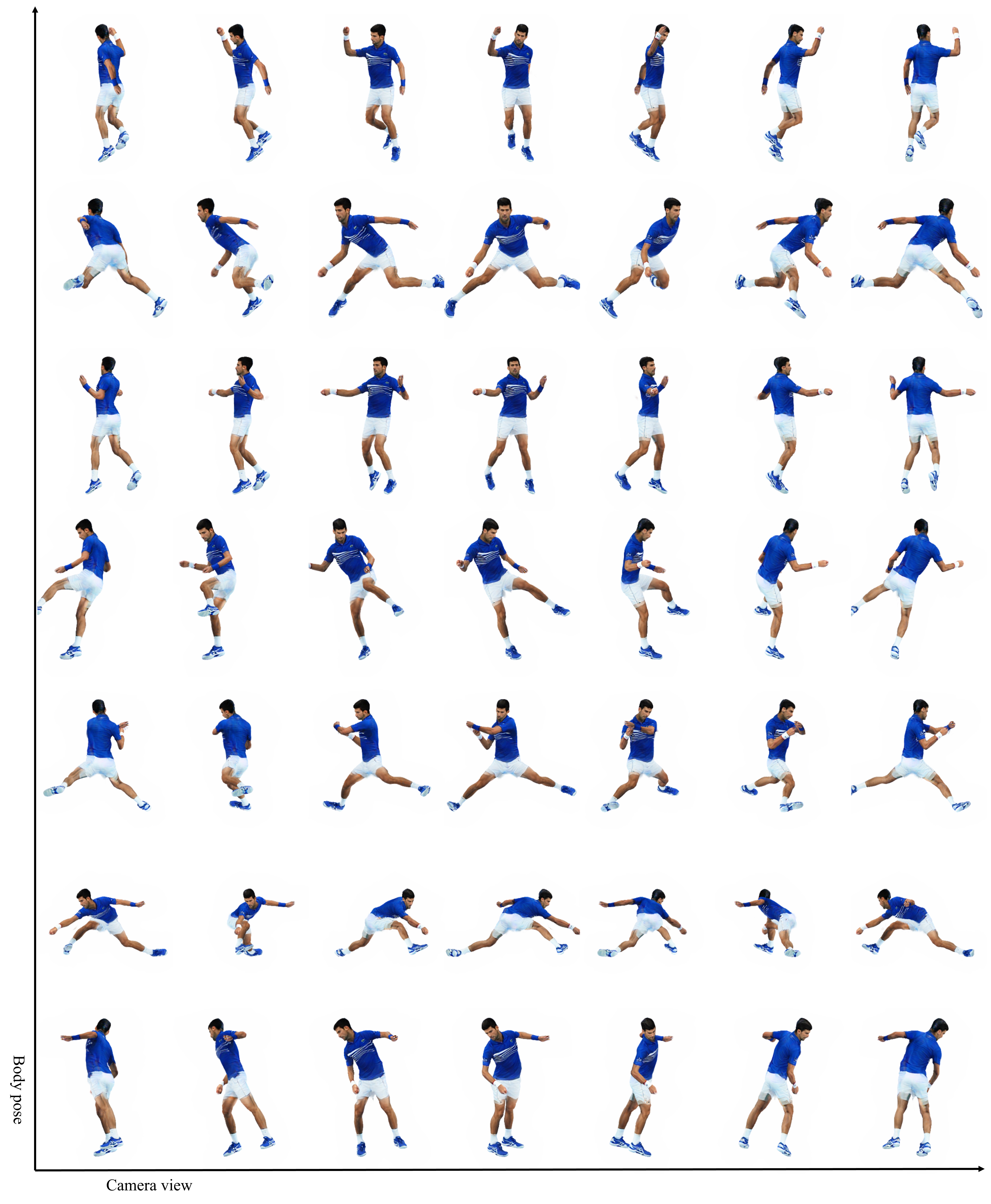}
  \caption{Visualization of the (body pose, camera view) plane of the reconstructed space of Novak Djokovic.}
  \label{fig:djoko_pos_view_plane}
\end{figure*}

\begin{figure*}
  \centering
  \includegraphics[width=\textwidth]{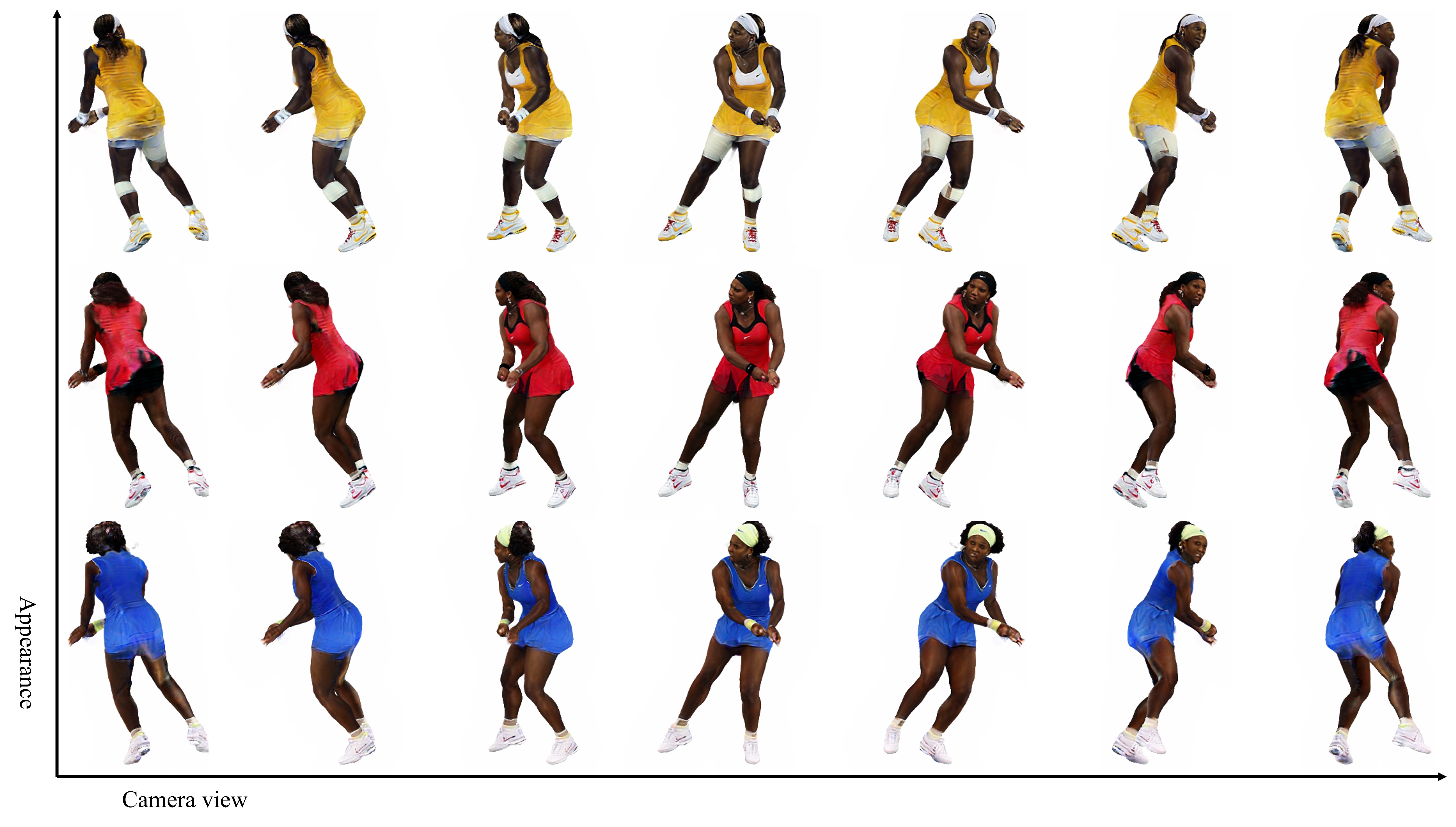}
  \caption{Visualization of the (appearance, camera view) plane of the reconstructed space of Serena Willaims.}
  \label{fig:serena_view_appearance_plane}
\end{figure*}

\begin{figure*}
  \centering
  \includegraphics[width=\textwidth]{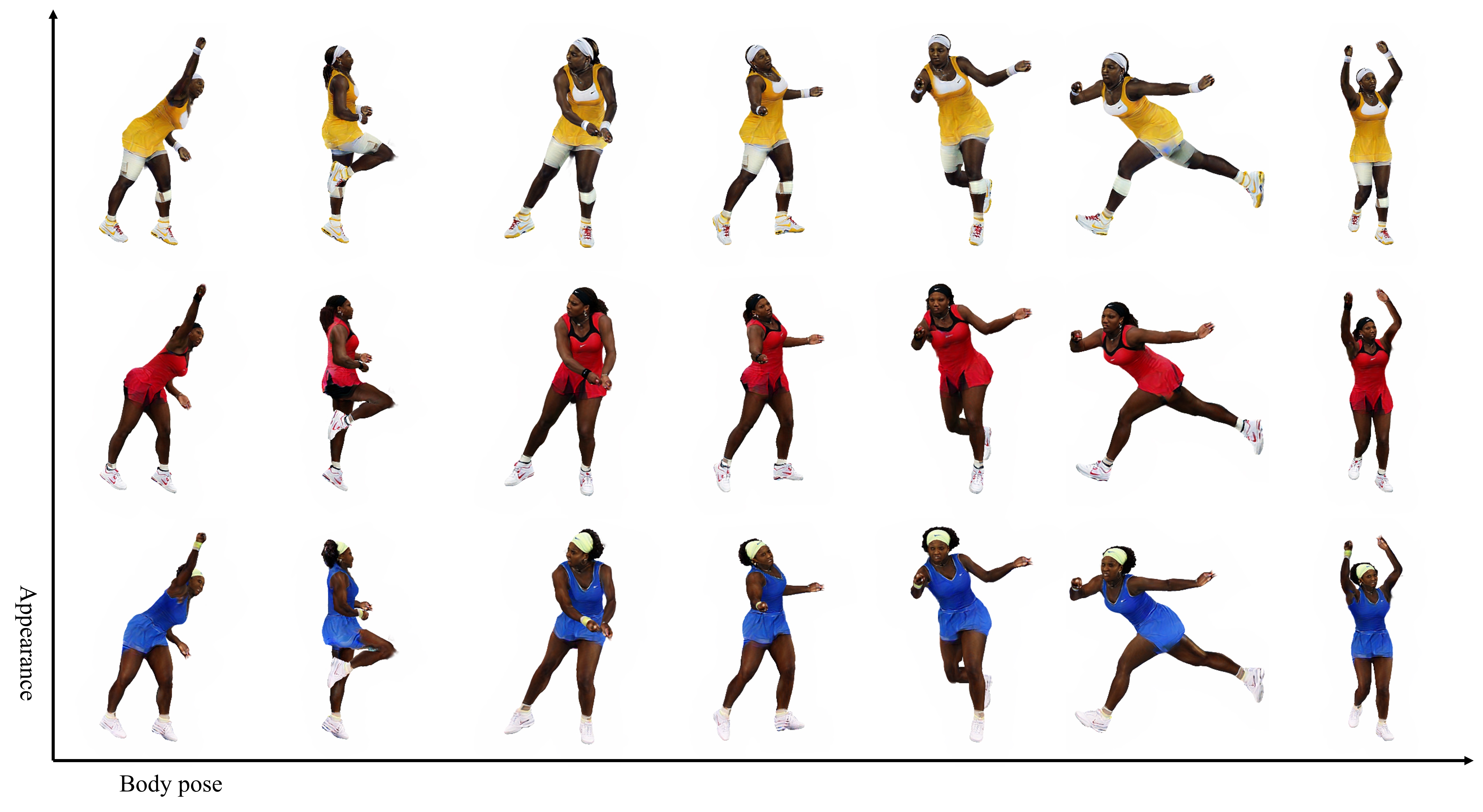}
  \caption{Visualization of the (appearance, body pose) plane of the reconstructed space of Serena Williams. }
  \label{fig:serena_pose_appearance_plane}
\end{figure*}

\begin{figure*}
  \centering
  \includegraphics[width=\textwidth]{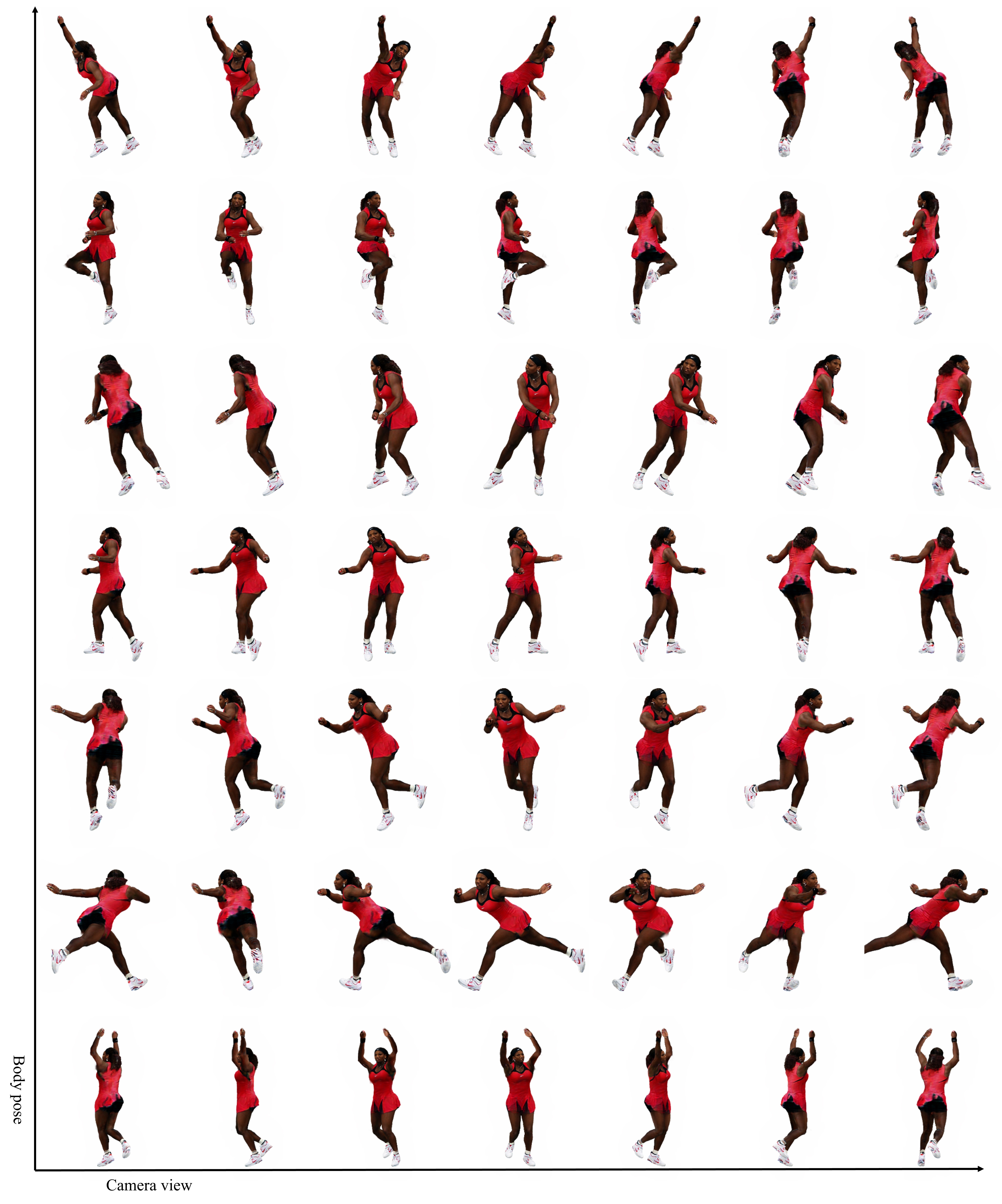}
  \caption{Visualization of the (body pose, camera view) plane of the reconstructed space of Serena Willaims.}
  \label{fig:serena_pos_view_plane}
\end{figure*}

\begin{figure*}
  \centering
  \includegraphics[width=\textwidth]{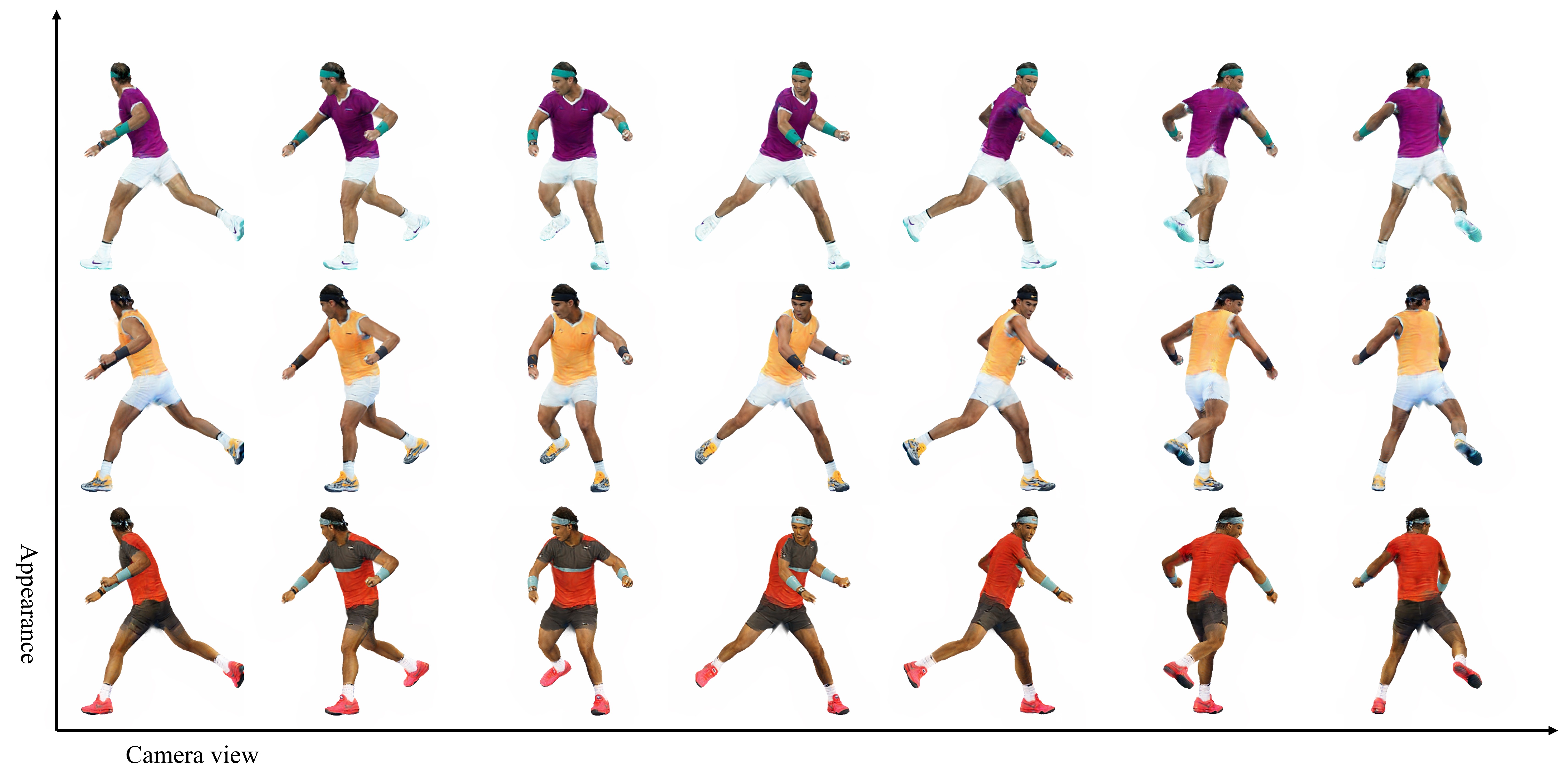}
  \caption{Visualization of the (appearance, camera view) plane of the reconstructed space of Rafael Nadal.}
  \label{fig:nadal_view_appearance_plane}
\end{figure*}

\begin{figure*}
  \centering
  \includegraphics[width=\textwidth]{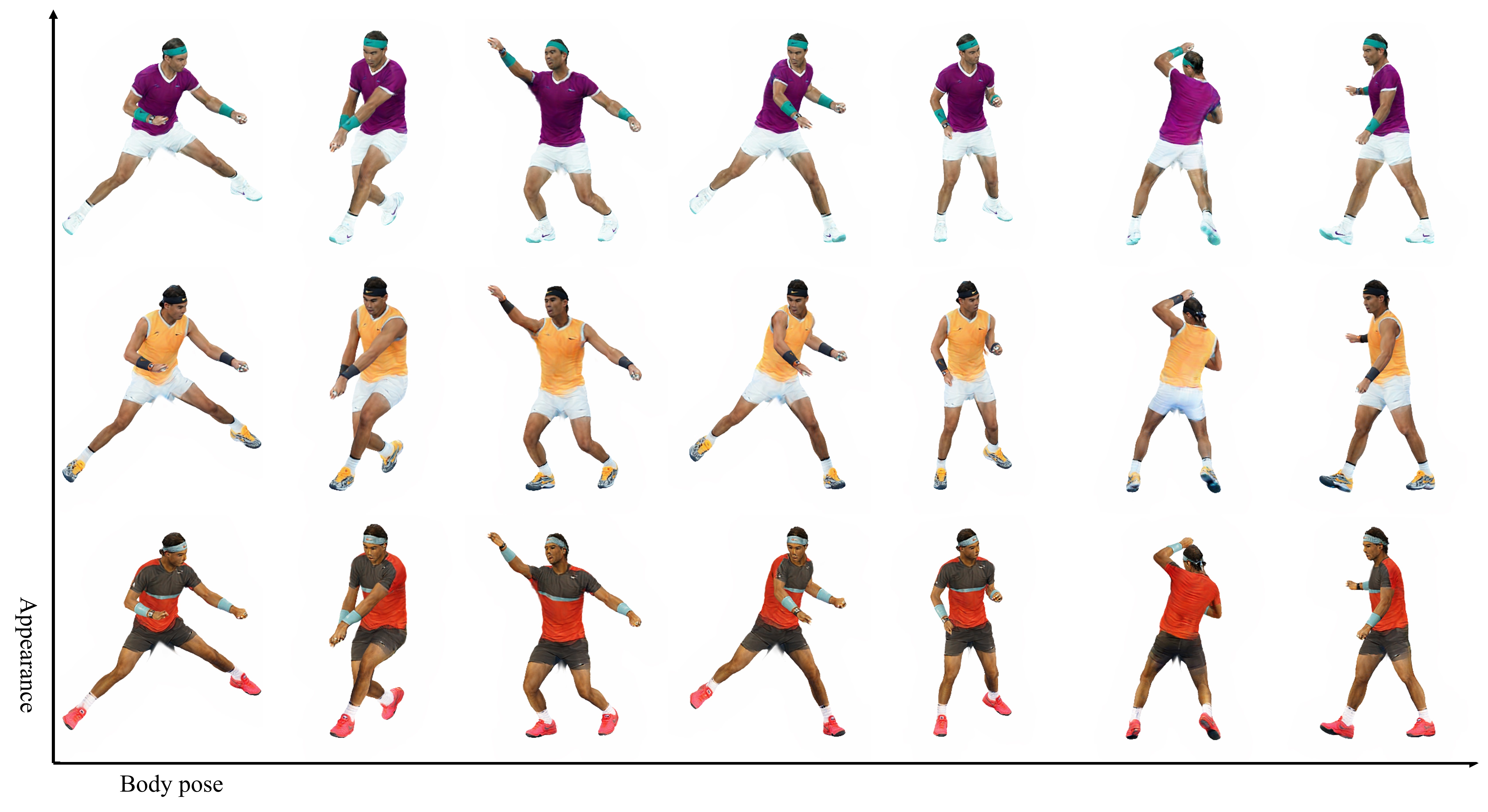}
  \caption{Visualization of the (appearance, body pose) plane of the reconstructed space of Rafael Nadal. }
  \label{fig:nadal_pose_appearance_plane}
\end{figure*}

\begin{figure*}
  \centering
  \includegraphics[width=\textwidth]{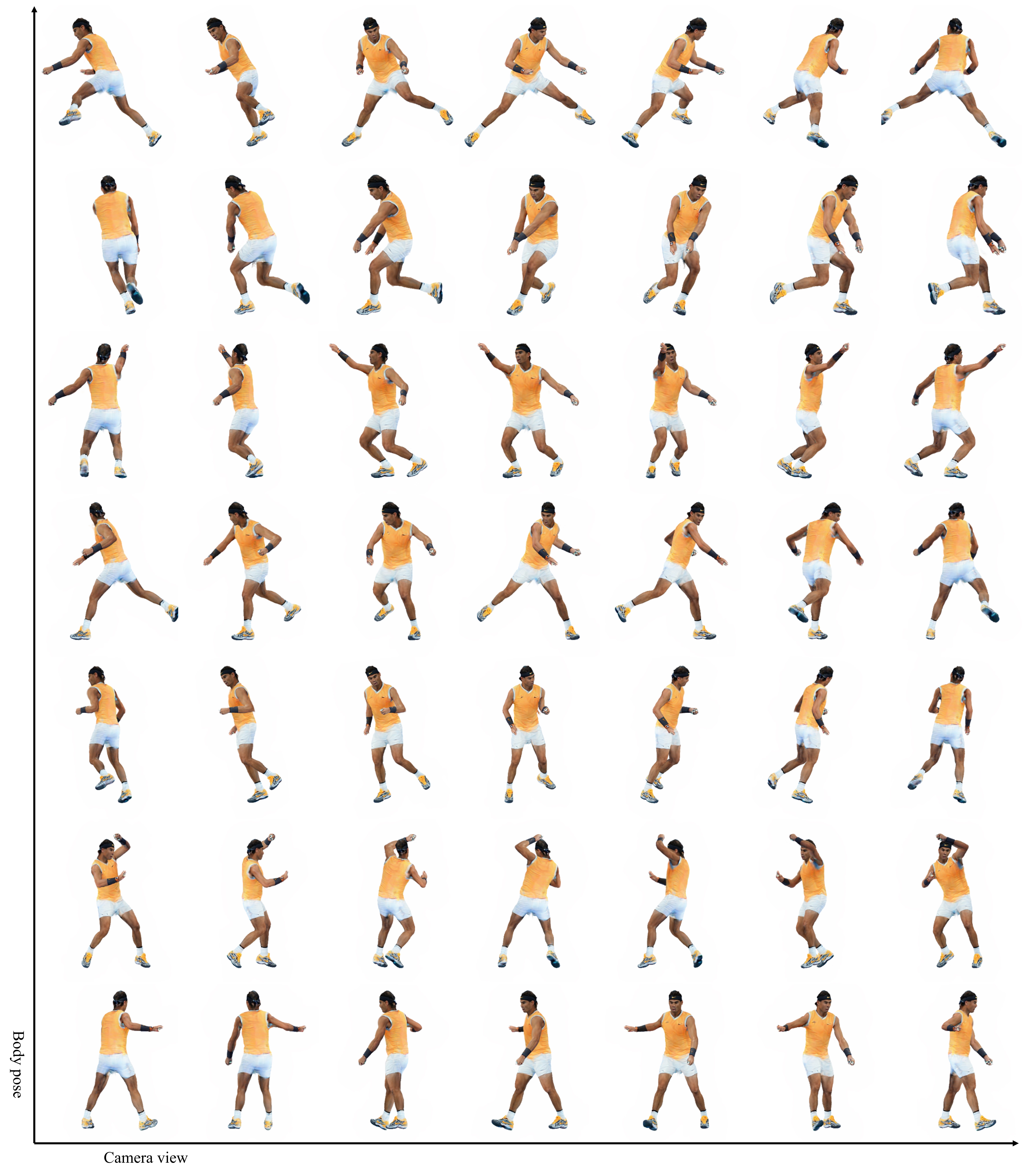}
  \caption{Visualization of the (body pose, camera view) plane of the reconstructed space of Rafael Nadal.}
  \label{fig:nadal_pos_view_plane}
\end{figure*}

Additionally, we show visualizations of the rebuilt personalized spaces of the other 3 persons, Novak Djokovic in Fig. \ref{fig:djoko_view_appearance_plane}, \ref{fig:djoko_pose_appearance_plane} and \ref{fig:djoko_pos_view_plane}, Serena Willams in Fig. \ref{fig:serena_view_appearance_plane}, \ref{fig:serena_pose_appearance_plane} and \ref{fig:serena_pos_view_plane}, and Rafael Nadal in Fig.  \ref{fig:nadal_view_appearance_plane}, \ref{fig:nadal_pose_appearance_plane} and \ref{fig:nadal_pos_view_plane}.





\end{document}